\documentclass[letterpaper, 10 pt, conference]{ieeeconf}  % Comment this line out if you need a4paper

\IEEEoverridecommandlockouts                              % This command is only needed if 
\usepackage{cite}
\usepackage{amsmath}
\usepackage{amssymb}
\usepackage{array}
\usepackage{subfiles} 
\usepackage{algorithm}
\usepackage{subfigure}
\usepackage{algorithm}
\usepackage{algpseudocode}
\usepackage{graphicx}
\usepackage{textcomp}
\usepackage{caption}

\newcommand{\argmax}{\operatornamewithlimits{argmax}}
\newcommand{\argmin}{\operatornamewithlimits{argmin}}

\DeclareMathOperator*{\Expec}{\mathbb{E}}

\title{\LARGE \bf
	A Distributed ADMM Approach to Non-Myopic Path Planning for Multi-Target Tracking
}

\author{Soon-Seo~Park\authorrefmark{1}, Youngjae Min\authorrefmark{1}, Jung-Su~Ha, Doo-Hyun~Cho and Han-Lim Choi\authorrefmark{2}% <-this % stops a space
	\thanks{{{Soon-Seo~Park and Han-Lim~Choi are with the Department of Aerospace Engineering \& KI for Robotics, KAIST. Youngjae Min is with School of Electrical Engineering, KAIST. Jung-Su~Ha is with Machine Learning and Robotics Lab, University of Stuttgart and Max Planck Institute for Intelligent Systems. Doo-Hyun~Cho is with Mechatronics R\&D center, Samsung Electronics.}}}    
	\thanks{{{\authorrefmark{1}Equal contribution. \authorrefmark{2}Corresponding author: hanlimc@kaist.ac.kr}}}    
}

\begin{document}

	\maketitle
	\thispagestyle{empty}
	\pagestyle{empty}

\begin{abstract}
	This paper investigates non-myopic path planning of mobile sensors for multi-target tracking.
	Such problem has posed a high computational complexity issue and/or the necessity of high-level decision making.
	Existing works tackle these issues by heuristically assigning targets to each sensing agent and solving the split problem for each agent.
	However, such heuristic methods reduce the target estimation performance in the absence of considering the changes of target state estimation along time.
	In this work, we detour the task-assignment problem by reformulating the general non-myopic planning problem to a distributed optimization problem with respect to targets. By combining alternating direction method of multipliers (ADMM) and local trajectory optimization method, we solve the problem and induce consensus (i.e., high-level decisions) automatically among the targets.
	In addition, we propose a modified receding-horizon control (RHC) scheme and edge-cutting method for efficient real-time operation. The proposed algorithm is validated through simulations in various scenarios.
\end{abstract}

	\maketitle
	
\section{Introduction} \label{intro}

Technological advances in sensor networks have enabled many applications such as monitoring pollution/atmospheric phenomena, reconnaissance and surveillance missions in both defense and civilian areas, traffic monitoring, forest fire localization and wildlife tracking \cite{dunbabin2012robots,yang2018optimal}.
These sensor networks are based on two types of sensors: static and mobile sensors. It is widely demonstrated that mobile sensors offer distinctive advantages over static ones in terms of the quality of sensing and estimation \cite{chhetri2006nonmyopic,hoffmann2010mobile,ponda2008trajectory,miller2009pomdp,oshman1999optimization,oh2013coordinated,farmani2017scalable,skoglar2007uav, jiang2018online}, coverage area, data offloading \cite{cheng2018uav} and adaptability to dynamic environments \cite{ragi2013uav}.
However, for mobile sensor networks, it is crucial to efficiently operate sensing agents to maximize information gain about target systems while satisfying constraints on entire systems.
This problem is described as a sensor planning problem that determines the future utilization of sensing agents given the current states of resources and environments.

One typical approach for the sensor planning problem is making myopic plans covering one or few time steps to reduce the complexity of the problem.
%Depending on the time span of the future plan, algorithms solving the sensor planning problem fall into two categories: myopic and non-myopic.
For instance, Hoffmann and Tomlin \cite{hoffmann2010mobile} proposed a single step mobile sensor planning algorithm for single target tracking based on an information-theoretic metric. 
They plan the one step ahead control inputs of the sensing agents to maximize the mutual information between the sensor measurements and the estimated target state.
%a particle filter for the purpose of estimating the state of the target more precisely when the stochastic dynamics of the target is non-Gaussian and nonlinear, and it requires discretization of the control inputs. Accordingly, it is difficult to use in multi-step ahead planning.
In multi-target tracking tasks, most existing works adopt the two-phase approach in which they cluster and assign targets to each sensing agent and, then, solve the split problem. \cite{oh2013coordinated,farmani2017scalable}.
However, such myopic planning algorithms are vulnerable to bad local optima and show poor performances for sensor networks with sensing holes \cite{liu2003multi,lee2018nonmyopic}. Their optimization over the near future cannot reflect long-term goals such as taking targets out of sensing holes by changing the positions and directions of sensors.
Thus, we focus on non-myopic planning algorithms which explore multiple time steps.

%The path planning of mobile sensors aims to optimize a sequence of waypoints/control inputs of sensor platforms to reduce the uncertainty of the target states.
%In this problem, not only the dynamics of the sensor platform but disturbances such as wind can also be considered for planning.
There are several existing approaches to non-myopic sensor planning problems.
One natural way is to extend sensor scheduling methods \cite{williams2007approximate,choi2013outer,lee2018potential} for static sensor networks.
As the methods select operating sensors among finite candidates, control inputs of the mobile sensors are discretized and the same optimization methods are applied for the finite optimization variables \cite{chhetri2006nonmyopic,lee2018nonmyopic}.
%Decision variables, such as selecting the sensor in the sensor scheduling problem, can be treated in the same way as variables such as the control input of the mobile sensor \cite{chhetri2006nonmyopic,lee2018potentialaccess}.
However, this approach suffers from a critical computational issue that the time complexity grows exponentially as the planning horizon increases.
%and requires an algorithm such as branch-and-bound based pruning \cite{chhetri2006nonmyopic}.
%Furthermore, because the control input has to be discretized, the dimension of the problem increases rapidly for sophisticated control of the sensor platform.
Another approach is the sampling-based planning. 
Information-rich Rapidly-exploring Random Tree (IRRT) algorithm~\cite{levine2010information} connects random sample points on the sensing space so that the sensor measurements along the connected paths maximize the information on a single target. Luders et al. \cite{luders2011information} apply this IRRT algorithm in a two-phase approach so that it tracks multiple targets. However, these sampling-based works do not guarantee any optimality in their solutions \cite{karaman2011sampling}.

A better candidate for non-myopic planning of mobile sensors is local trajectory optimization method.
It searches for a solution to the optimization problem near the given initial guess using gradient descent algorithms.
This gradient-based trajectory optimization has been successfully applied in many single-target tracking applications by employing analytically differentiable filter models on target estimation \cite{ponda2008trajectory,miller2009pomdp,oshman1999optimization,ragi2013uav,skoglar2007uav}. 
%Such formulations have a great advantage in handling the dynamic constraints of the sensor platform and/or environment.
%For instance, a trajectory planning method that takes into account the effects of wind on the sensor platform and collision-avoidance was proposed \cite{ragi2013uav}.
%Oshman et al. investigated optimal trajectory planning subject to state constraints imposed on the sensor platform \cite{oshman1999optimization}. 
%Ponda et al. \cite{ponda2008trajectory} used an FIM to quantify the amount of information acquired by sensor measurements and analyzed an optimality criterion for a bearings-only target localization problem.
It shows much faster computation time than the discrete optimization methods and guarantees local optimality unlike the sampling-based methods.
However, in multi-target tracking problems, local trajectory optimization methods can be easily trapped in a bad local optimum unless a proper initial guess is given.

%The targets are assumed to be static, and data mining tools such as the K-mean clustering algorithm or density-based spatial clustering of applications with noise (DBSCAN) \cite{ester1996density} are used to form clusters based on the distance separating the targets.
%Clusters with large uncertainties or large numbers of targets are designated with higher priority over others for mobile sensor assignments.
%Tasks are assigned to mobile sensors based on the priority of the clusters and the distance between the mobile sensors and the center of the clusters.
%Then, one or two step trajectory planning is performed \cite{oh2013coordinated,farmani2017scalable}.

In this paper, we investigate non-myopic path planning of mobile sensors for multi-target tracking problems. 
There is a key challenge to employ local trajectory optimization method in multi-target tracking problems; to make a good initial guess, we need a high-level decision-making process that 
%before trajectory optimization. Specifically, decision-making should
determines which mobile sensors should track which targets and in what order the targets should be tracked.
This problem is also referred to as the task-assignment problem.
Unfortunately, the problem is hard to be formulated to reflect the mobility of the sensors and expected changes in the states of the sensors and the targets.
Instead, existing works, while they are myopic, with two-phase approach use heuristics for the formulation \cite{oh2013coordinated,farmani2017scalable}, which degrade the quality of initial guess and, thus, the target estimation performance.
Therefore, the objective of this study is to solve the non-myopic path planning problem for multi-target tracking without heuristic task-assignment.

%Existing local trajectory optimization-based algorithms focus on the non-myopic planning of mobile sensors rather than solving task-assignment problems \cite{ponda2008trajectory,miller2009pomdp,oshman1999optimization,ragi2013uav,skoglar2007uav}. In contrast, two-phase approaches were developed for the purpose of task-assignment rather than non-myopic trajectory planning \cite{oh2013coordinated,farmani2017scalable}.
%This work aims to bridge the gap between preliminary work on non-myopic trajectory planning \cite{ponda2008trajectory,miller2009pomdp,oshman1999optimization,ragi2013uav,skoglar2007uav} and task-assignment \cite{oh2013coordinated,farmani2017scalable,luders2011information}.
%In the trajectory planning problem for multiple target tracking, it is necessary to determine which mobile sensors should track which targets, in what order the targets should be tracked, and how much time should be devoted to each target.
%To the best of our knowledge, no attempt has been made to handle both non-myopic trajectory planning and task-assignment in multi-target tracking problems simultaneously while considering the constraints of the sensor platform mobility and changes in target states over time.

The main contribution of this paper is to provide a new practical algorithm without heuristic task-assignment on the non-myopic path planning problem for multi-target tracking. More specifically, we\\
\vspace{-0.4cm}
\begin{enumerate}
%	[label=(\roman*)]
	\item reformulate the general non-myopic path planning problem for multi-target tracking to a distributed optimization problem with respect to targets,
	\item solve the transformed problem by combining ADMM and local trajectory optimization method, and
	\item propose a modified receding-horizon control (RHC) scheme and edge-cutting method for efficient real-time operation
\end{enumerate}
Our algorithm circumvents the task-assignment problem by solving subproblems for each target, which have only single task, and inducing consensus among those subproblems automatically.
Consequently, our distributed framework integrates the decision-making process into the trajectory optimization process so that they are solved simultaneously without the aid of heuristic task-assignment algorithms.

%Second, we propose a receding-horizon control (RHC) scheme and an edge-cutting method for real-time implementation of the proposed algorithm. Our RHC scheme makes it possible to plan the future trajectory while taking into account the computation time of the trajectory optimization algorithm, which includes motion estimations of the targets. The edge-cutting method serves to reduce the dimensions of the subproblems during the optimization process. If the tuning parameters are set appropriately, the edge-cutting method reduces the computation time of the algorithm while yielding the same result as not applied.

The alternating direction method of multipliers (ADMM) \cite{boyd2011distributed,mota2013d,zhang2014asynchronous} has been also effectively applied to other problems in robotic control and sensor scheduling \cite{bento2013message,ong2015cooperative,derbinsky2013improved,mordatch2014combining,liu2014optimal}. 
For instance, Bento et al. \cite{bento2013message} adopted ADMM to multi-agent trajectory planning for collision-avoidance. ADMM is used to incorporate various constraints, such as energy minimization, with the planning objective.
%In particular, they proposed an improved version of ADMM via message-passing for fast convergence of a solution.
Ong et al. \cite{ong2015cooperative} also developed a message-passing algorithm based on ADMM to solve a collision-avoidance problem with communication constraints.
Mordatch et al. \cite{mordatch2014combining} combined trajectory optimization and global policy (neural network) learning using ADMM in a robotic control problem.
%The joint optimization problem enables the trajectory optimizer to act as a teacher for neural network training rather than as a demonstrator.
Liu et al.\cite{liu2014optimal} used ADMM to solve an optimization problem with cardinality functions in scheduling static sensor networks for state estimation of dynamical systems.
%The problem is formulated to strike a balance between estimation accuracy and the total number of sensor activations over one period.
%They use ADMM to solve the optimization problem, including cardinality functions. 
Like these works, we formulate a problem on which ADMM can work effectively and successfully solve the problem.

The rest of this paper is organized as follows. 
In Section \ref{sec:preliminary}, we define the general non-myopic path planning problem for multi-target tracking.
Then, we reformulate the original problem as a distributed trajectory optimization problem in Section \ref{sec:Transformation}. 
Section \ref{sec:mainAlg} presents the details of our algorithm including the modified RHC scheme.
%; we describe the mathematical formulation of the distributed trajectory optimization algorithm. 
In Section \ref{sec:complexity}, computational complexity of the proposed algorithm is analyzed and the edge-cutting method is introduced to reduce the complexity.
%   Finally, the proposed algorithms are verified and some intuitions are given by numerical simulations in Section \ref{sec:simulation}.
Finally, the simulation results of the proposed algorithm are given in Section \ref{sec:simulation}.

\section{Problem Formulation} \label{sec:preliminary}

%We begin by defining the trajectory optimization problem of mobile sensors to enhance the tracking performance. 
%The goal of a target tracking problem is to estimate the kinematic states of targets with a finite set of measurements, which naturally fits into Bayesian state estimation formulation (i.e., Bayesian filtering).
%Bayesian filtering formulation provides us with a measure for evaluation of the tracking performance. 
%This measure represents the uncertainty of the target states, and the location of the mobile sensors should be planned to reduce it, that is, to enhance the tracking performance, over a specified time horizon.
%First, we present the mobile sensor and target models for the sake of concreteness. These models are used to define the trajectory planning problem of mobile sensors.
%Then, we formulate the distributed trajectory optimization for the multi-target tracking problem.

\subsection{Non-Myopic Path planning for multi-target tracking}
We consider the problem of non-myopically planning multiple mobile sensors to efficiently track multiple targets. We formulate this problem based on the known dynamics of targets and sensors, as well as the sensing model. From these models, we plan an optimal path that each sensor takes successive measurements to estimate target states efficiently. 

In this work, targets are assumed to follow linear stochastic dynamics which is a common assumption in the literature \cite{lee2018nonmyopic,ragi2013uav,chhetri2006nonmyopic,ponda2008trajectory}. Let the index set of targets be $\mathcal{T}=\{1,2,\cdots,M\}$,
%		and let $\mathcal{Z}\subset\mathbb{R}^o$ be the space of all possible sensor measurements, $\textbf{z}$, that the mobile sensor may receive. 
where the number of targets, $M$, is known and fixed. 
%we can ignore the joint distribution and use one tracking filter for each target \cite{skoglar2007uav}. 
The $j$-th ($j\in\mathcal{T}$) target's state, $\textbf{x}^{(j)}_t$, is updated as:	
\begin{equation} \label{eq:linear_stochastic}
\textbf{x}_{t+1}^{(j)} = A_t^{(j)}\textbf{x}_{t}^{(j)}+\omega_t^{(j)},
\end{equation}
where $\omega_t^{(j)}\sim \mathcal{N}(0,\Sigma_{w,t}^{(j)})$ is the Gaussian random noise independent of 
other targets and measurements. $A_t^{(j)}$ and $\Sigma_{w,t}^{(j)}$ are the transition matrices and noise covariance matrices of the $j$-th target, respectively.

For sensor dynamics, we employ a general model, $f^{(i)}$, for each sensor. Let the index set of mobile sensors be $\mathcal{A}=\{1,2,\cdots,N\}$. The dynamics for the $i$-th ($i\in\mathcal{A}$) sensor's state, $p_t^{(i)}$, is expressed as:

\begin{equation}
p_{t+1}^{(i)} = f^{(i)}(p_t^{(i)},u_t^{(i)}),
\end{equation}
where $u_t^{(i)}$ is the control input at time $t$.
%Let $\mathcal{Z}\subset\mathbb{R}^o$ be the space of all possible sensor measurements that the mobile sensors receive. 
Here, each mobile sensor is assumed to be carried by one fixed-wing UAV for simplicity.
\footnote{ In this paper, we use the terms `mobile sensor' and `UAV' interchangeably.
	For simplicity, we assume that each UAV has one sensor on it. In general, some platforms may carry more than one sensor, and the sensors may be heterogeneous. In such cases, the sensors on the same vehicle belong to the same dynamics.}

Denoting the measurement taken by the $i$-th UAV against the target $j$ at time $t$ as $z_t^{(i,j)}$, the sensor measurement model can be expressed as follows:
%		$i^{\text{th}}$
\begin{equation}
z_t^{(i,j)} = h^{(i)}(\textbf{x}_t^{(j)},p_t^{(i)})+v_t^{(i)}, 
\end{equation}
where $v_t^{(i)}\sim \mathcal{N}(0,\Sigma_{v,t}^{(i)})$ is the Gaussian random noise, independent of other measurements and targets' motion noise, $w_t^{(j)}$.

These models are incorporated to form an objective function, $\mathbb{U}^{(j)}$, which indicates the uncertainty of the target state estimation. The goal of the general target tracking problem is to estimate target states accurately, which corresponds to minimizing the uncertainty. Since we are solving a non-myopic path planning problem, the optimization variables are a sequence of states/control inputs of mobile sensors over certain future time steps, $T$. Thus, in general, non-myopic path planning for multi-target tracking can be formulated as \cite{lee2018nonmyopic,ragi2013uav,chhetri2006nonmyopic}:

%The goal of general non-myopic path planning for multi-target tracking is to optimize a sequence of states/control inputs of mobile sensors over a certain future time step $T$ to minimize the uncertainty $ \mathbb{U}^{(j)}$ of the targets \cite{lee2018nonmyopic,ragi2013uav,chhetri2006nonmyopic}.
%Given a specific initial state and target state, one can find an optimal trajectory $\big[\bar{\textbf{P}},\bar{\textbf{U}} \big]$ that satisfies the dynamics constraints:

%\RE{[ref extension uncertainty, objective function]
%The purpose of the problem is finding the optimal sequence of control inputs, U, and the resulted waypoints, P, which minimize.
%General non-myopic path planning problem can be formulated as follows:}
%Thus, the planning objective is to optimize a sequence of states/control inputs of mobile sensors over a certain future time step $T$ to minimize the uncertainty of the targets.
%
%Given a specific initial state and target uncertainty, one can find an optimal trajectory $\big[\bar{\textbf{P}},\bar{\textbf{U}} \big]$ that satisfies the dynamics constraints:
\begin{equation}
\begin{split}
\big[\bar{\textbf{P}},\bar{\textbf{U}} \big] = &\argmin_{\textbf{p},\textbf{u}} 
\sum_{t=0}^{T} \Bigg[  \sum_{j=1}^{M} \mathbb{U}^{(j)}(\textbf{p}_{t})+\textbf{u}_t^{\top} R \textbf{u}_t + \eta(\textbf{p}_{t})\Bigg]:\\
&\textbf{p}_{t+1} = \textbf{f}(\textbf{p}_t,\textbf{u}_t) \quad and  
\quad \textbf{p}_0 = \textbf{p}_{init},
\end{split}	\label{eq:original_cost}
\end{equation}		
where 
\begin{equation*} %\small
\begin{split}
\textbf{f}(\textbf{p}_{t},\textbf{u}_{t}) = \Big[{f^{(1)}(p_{t}^{(1)},u_{t}^{(1)})}^{\top},\cdots,{f^{(N)}(p_{t}^{(N)},u_{t}^{(N)}}^{\top})\Big]^{\top},\\
\textbf{p}_t = \Big[{p_t^{(1)}}^{\top},\cdots,{p_t^{(N)}}^{\top} \Big]^{\top},
\textbf{u}_t = \Big[{u_t^{(1)}}^{\top},\cdots,{u_t^{(N)}}^{\top} \big]^{\top}.
\end{split} 
\end{equation*}
Here, the optimal trajectory, $\big[\bar{\textbf{P}},\bar{\textbf{U}} \big]$, satisfies the dynamics constraint and the given initial condition.
$\mathbb{U}^{(j)}(\textbf{p}_{t})$ is the uncertainty of the $j$-th target at time  $t$, $\textbf{u}_t \textit{R}_t \textbf{u}_t^T$ penalizes the control effort of the mobile sensors along the trajectory, and $\eta(\textbf{X}_t)$ is an additional constraint, such as collision-avoidance, related to the states of the mobile sensors. $R$ is a weighting matrix. 
%The problem defined in (\ref{eq:original_cost}) is a general trajectory optimization problem for a stochastic system.

%\section{Uncertainty Measure} \label{sec:uncertainty}

%To complete the problem formulation, we introduce 

\subsection{Target State Estimation} \label{subsec:estimation}
The target states are estimated through the extended Kalman filter (EKF), which is widely used for state estimation of non-linear systems \cite{chen2003bayesian}. Note that any other non-linear Gaussian filters, such as the unscented Kalman filter, can be incorporated instead.
The targets are assumed to have independent motions and be measured independently via data association methods\cite{bar2011tracking}. Then, their joint distribution could be factorized and a single tracking filter is used for each target \cite{skoglar2007uav}.

For each target, its belief states, $(\hat{\textbf{x}}_{t}^{(j)},\Sigma_t^{(j)})$, are updated by 
(for simplicity, the superscript $(j)$ is temporarily dropped): 

\vspace{-0.3cm}
\begin{align}
%		\hat{\textbf{x}}_{t+1} &= \bar{\textbf{x}}_{t+1} +
\hat{\textbf{x}}_{t+1} &= A_t \hat{\textbf{x}}_{t}+
K_{t}(\textbf{z}_{t+1}-\textbf{h}(\hat{\textbf{x}}_{t+1},\textbf{p}_{t+1})), \label{eq:ekf_state}\\
\Sigma_{t+1} &= (I-K_t H_t)\bar{\Sigma}_{t}, \label{eq:ekf_cov}
\end{align}
%		\end{equation}		
where 
\begin{align*}
\bar{\Sigma}_{t} &= A_t \Sigma_{t} {A_t}^{T}+\Sigma_{w,t}, \\
K_t &= \bar{\Sigma}_{t} {H_t}^T (H_t \bar{\Sigma}_{t} {H_t}^T + \boldsymbol{\Sigma}_{v,t})^{-1},\\ 
\boldsymbol{\Sigma}_{v,t} &= diag([\Sigma_{v,t}^{(1)},\cdots,\Sigma_{v,t}^{(N)}]), \\
H_t &= \dfrac{\partial \textbf{h}(\textbf{x}_t, \textbf{p}_t)}{\partial \textbf{x}_{t}} 
\big|_{\hat{\textbf{x}}_{t},\textbf{p}_{t}}, \\
\textbf{h} &= [h^{(1)}(\textbf{x}_t, p_t),\cdots,h^{(N)}(\textbf{x}_t, p_t)]^{\top},\\
\textbf{p}_t &= \Big[{p_t^{(1)}}^{\top},\cdots,{p_t^{(N)}}^{\top}\Big]^{\top},\\
\textbf{z}_t &= \Big[{z_t^{(1,j)}}^{\top},\cdots,{z_t^{(N,j)}}^{\top}\Big]^{\top}
. 
\end{align*}
Equations (\ref{eq:ekf_state}) and (\ref{eq:ekf_cov}) define the belief dynamics. The second term in (\ref{eq:ekf_state}), called the innovation term, depends on the measurements $\textbf{z}_{t+1}$.
Equation (\ref{eq:ekf_cov}) evolves given the current covariance of the target, regardless of the measurement. 
%This represents the uncertainty of the target states and can be used as a measure to determine the sensing position of the mobile sensors [ref].	
%		Since the measurement is unknown in advance, the belief dynamics are stochastic.
%the belief dynamics defined within the EKF update is stochastic;

Our optimization problem in \eqref{eq:original_cost} evaluates the target state estimations in future time steps. 
Since future measurements are unknown in advance, we treat the innovation term in \eqref{eq:ekf_state} as stochastic, which follows $\mathcal{N}(0,K_t H_t \bar{\Sigma}_{t})$ \cite{van2012motion}. 
%		???? if we neglect this term, it return to maximum likelihood assumption?[ref].
Defining the belief state of the $j$-th target as $\textbf{b}_t^{(j)}= \big[{\hat{\textbf{x}}_t^{(j){\top}}} , vec[\Sigma_t^{(j)}]^{\top}  \big]^{\top}$ ($vec[\cdot]$ is a vector with upper triangular portion of the input matrix since $\Sigma_t^{(j)}$ is symmetric), the stochastic belief dynamics of the target is given by:
\begin{equation}
\textbf{b}_{t+1} = \textbf{g}(\textbf{b}_{t},\textbf{p}_t) +  \boldsymbol{\psi}_t, \quad  \boldsymbol{\psi}_t \sim	\mathcal{N}(0, {\Psi}),
\label{eq:belief_dyn}
\end{equation} 
where
\begin{equation}
\begin{split}	
\textbf{g}(\textbf{b}_{t},\textbf{p}_t) &= 
\begin{bmatrix}
A_t \hat{\textbf{x}}_t \\ vec[(I-K_t H_t)\bar{\Sigma}_{t}]
\end{bmatrix},\\
{\Psi}(\textbf{b}_{t},\textbf{p}_t) &= Var[\textbf{b}_{t+1}] = 		
\begin{bmatrix}
K_t H_t\bar{\Sigma}_{t} & 0 \\ 0 & 0
\end{bmatrix}.				 
\end{split} 
\end{equation} 
%\RE{Here, if we assume maximum-likelihood observations \cite{platt2010belief}, the stochastic term is removed from (\ref{eq:belief_dyn}), which makes belief propagation deterministic.} 
%		Discussions on the effect of maximum-likelihood observations can be found in previous work \cite{van2012motion}.

\subsection{Uncertainty Measure}
The uncertainty term in \eqref{eq:original_cost} represents the inaccuracy of target state estimation. For this measure, we use
\begin{equation}
\mathbb{U}^{(j)}(\textbf{p}_{t}) = \Expec_{\textbf{z}_t^{(j)}} \Big[ tr\big(\Sigma_t^{(j)}(\textbf{p}_{t})\big) \Big].
\end{equation}
The trace of the covariance matrix of target state estimation physically means the average variance of the estimation error and is one of the widely used metrics \cite{levine2010information,ponda2008trajectory,farmani2017scalable,ragi2013uav}. This measure is in a quadratic scale for physical distance error. 
%, so we can interpret  four times lower uncertainty as approximately 2 times lower distance error.
The expectation is taken since we are employing stochastic belief dynamics as in \eqref{eq:belief_dyn}. Note that any other differentiable metric can be incorporated instead.

%\RE{dependence on b}
%minimization of average uncertainty metric over the finite time horizon.

\section{Transformation to Distributed Trajectory Optimization Problem} \label{sec:Transformation}
\subsection{Challenge of Non-Myopic Problem} \label{subsec:challenge}
The non-myopic path planning problem in (\ref{eq:original_cost}) generally faces an abundance of local optima due to the non-convexity of its cost function.
This non-convexity primarily comes from the dependency of optimal sensor trajectories on high-level decision-making.
They significantly vary upon which target is assigned to which sensor and the order in which each sensor tracks the assigned targets.
Since most trajectory optimization methods find solutions around an initial guess, our problem with numerous local optima is very sensitive to initialization.
Therefore, we need to guide a solution by properly initializing the optimization variables.

Finding a good initial guess for the problem induces another hard problem, task-assignment. 
%requires to solving task-assignment problem.
Unfortunately, it is very difficult to construct the task-assignment problem in our setting to reflect the mobility of the sensors and expected changes in the states of the sensors and the targets.
%Constructing the task-assignment problem in our setting is very difficult because the cost function (\ref{eq:original_cost}) depends not only on time but also on the states of the mobile sensors and targets.
One feasible option is to heuristically assign tasks based on the initial states of targets and sensors utilizing the traveling salesman problem (TSP) \cite{reinelt1991tsplib}, CBBA \cite{choi2009consensus,luders2011information}, and/or clustering  \cite{oh2013coordinated,farmani2017scalable}.
%approximate future values of cost function from the current time based on information of targets and find solutions by solving task-assignment problems such as the traveling salesman problem (TSP) \cite{reinelt1991tsplib}, CBBA \cite{choi2009consensus,luders2011information}, or clustering  \cite{oh2013coordinated,farmani2017scalable}.
However, it is still hard to consider the change of targets and sensors' states over time sufficiently, and the option may provide a bad local optimum.

%These approaches, however, can provide a local optimal solution even in the myopic trajectory planning problem due to approximate/heuristic cost, and are not suitable for non-myopic trajectory planning problems that requires consideration of the change of a target states and sensor platform mobility.

%Initialized optimization variables are called an initial guess.
%, and the initial
%guess can be interpreted as having a role similar to task-assignment in a multi-target tracking problem.
%As noted earlier
%\RE{Unfortunately, constructing a task-assignment problem is very difficult because the cost function (\ref{eq:original_cost}) depends not only on the time but also on the states of the mobile sensors and targets.}

%These approaches, however, can provide a local optimal solution even in the myopic trajectory planning problem due to approximate/heuristic cost, and are not suitable for non-myopic trajectory planning problems that requires consideration of the change of a target states and sensor platform mobility.} 

\subsection{Transformed Problem}

Instead of solving the hard problem, we detour the difficulty and reformulate the original problem (\ref{eq:original_cost}) into a distributed trajectory optimization problem. For each target, we solve:
\begin{equation}
\begin{split}
\big[\bar{\textbf{P}},\bar{\textbf{U}} \big] = &\argmin_{\textbf{u}}  \sum_{j=1}^{M} \textbf{J}_j(\textbf{P},\textbf{U}):\\
&\textbf{p}_{t+1} = \textbf{f}(\textbf{p}_t,\textbf{u}_t) \quad and 
\quad \textbf{p}_0 = \textbf{p}_{init}, 
\end{split} \label{eq:dist_cost}
\end{equation}
where 
\begin{equation*}
\begin{split}
\textbf{J}_j(\textbf{P},\textbf{U}) &= 
\sum_{t=0}^{T}  \mathbb{U}^{(j)}(\textbf{p}_{t}) + 
\dfrac{1}{M}\big(\textbf{u}_t^{\top} R_t \textbf{u}_t + \eta(\textbf{p}_{t})\big),\\
\textbf{P} &= [\textbf{p}_1,\cdots,\textbf{p}_T ], \quad \textbf{U}= [\textbf{u}_1,\cdots,\textbf{u}_T ].
\end{split}
\end{equation*}		
%The trajectory optimization problem for each target can be handled more easily than multiple target tracking problems because task-assignment is not required
Trajectory optimization for each target is much easier than that for multiple targets because task-assignment is not considered.
Also, it is not difficult to generate the initial guess for the problem of tracking a single target.
%Since the trajectory optimization results for each subproblem must be reflected in the optimization of each of the other subproblems and the results must converge to the same solution, subproblems need to be coupled.
%Based on the fact that trajectory optimization problems for one target can be handled more easily than multiple target tracking problems because task-assignment is not required,
%The trajectory optimization results for each subproblem must be reflected in the optimization of each of the other subproblems and the results must converge to the same solution.
By solving the trajectory optimization for each target and making consensus, this distributed optimization is solved through Alternating Direction Method of Multipliers (ADMM) \cite{boyd2011distributed,wei2012distributed}.

\subsubsection{Distributed Alternating Direction Method of Multipliers} \label{subsec:admm}
ADMM is an algorithm that efficiently optimizes objective functions composed of terms that each has efficient solution methods \cite{boyd2011distributed}.
%		ADMM \cite{boyd2011distributed} is an algorithm that can be implemented for large-scale convex optimization problem in a distributed manner. 
%		ADMM is a method that can be used to efficiently optimize objective functions composed of terms that each independently have efficient solution methods. 
For our purposes, we use a consensus and sharing optimization form of ADMM, also known as distributed ADMM \cite{wei2012distributed,mota2013d,zhang2014asynchronous}.
We consider the following minimization problem:
\begin{equation}
\begin{split}
\min \sum_{l=1}^{L} \textit{J}_l(a).
\end{split}
\end{equation}	
This problem can be rewritten with the local variable $a_l$ and common global variable $b$:
\begin{equation}
\begin{split}
&\min \sum_{l=1}^{L} \textit{J}_l(a_l): \enspace a_l - b = 0,\enspace  l=1,\cdots,L.
\end{split}
\end{equation}	
This is called the global consensus problem, as the constraint is that all the local variables should be equal (i.e., agree). Solving this problem is equivalent to optimizing the augmented Lagrangian:
%		\begin{equation}
%			\mathcal{L}({\{a_l\},b;\{\lambda_l\}}) = \sum_{l=1}^{L} {\big(\textit{J}_l(a_l)+\dfrac{\rho}{2} {\lVert{a_l-b+ \lambda_l}\rVert}^2 \big)},  
%		\end{equation}
\begin{equation}
\mathcal{L}({\lambda_l}) = \sum_{l=1}^{L} {\big(\textit{J}_l(a_l)+\dfrac{\rho}{2} {\lVert{a_l-b+ \lambda_l}\rVert}^2 \big)},  
\end{equation}		
where $\lambda_l$ is the Lagrange multipliers for ${a_l-b=0}$, and $\rho > 0$ is a penalty weight.
The ADMM algorithm updates the local variable $a_l$, the common global variable $b$ and the Lagrange multiplier $\lambda_l$ as follows.
%		The ADMM algorithm can be described as follows. 
For $k = 0,1,\cdots,$ we iteratively execute the following three steps:
\begin{equation} \label{admmeq}
\begin{split}
a_l^{k+1}&=\argmin_{a_l}\textit{J}_l(a_l)+\dfrac{\rho}{2} {\lVert{a_l-b^k+\lambda_l^k}\rVert}^2,
%\\ &\quad l=1,\cdots,L, 
\\
b^{k+1}&=\frac{1}{L} \sum_{l=1}^{L} (a_l^{k+1}+\lambda_l^k), \\
\lambda_l^{k+1}&=\lambda_l^k+\frac{\alpha}{\rho}(a_l^{k+1}-b^{k+1}), 
%\enspace l=1,\cdots,L ,
\end{split}
\end{equation}
where $\alpha>0$ is a step-size parameter. Here, note that the first and the last steps can be carried out independently for each $l=1,\cdots,L$.
Thus, we can run the algorithm in parallel depending on the implementation with multiple but no more than $L$ computational devices, where each device performs an update on its set of variables.	
%		If we had $L$ computational devices, we could assign each $a_l$ to a separate computational device. Then optimization over $a_l$ can be done in parallel. Subsequently, the results are communicated back to a master node which performs the global variable $b$ update and returns back the result to other worker nodes. 
The optimization results  $a_l$ are communicated back to a master node, which performs the global variable $b$ update and returns the result back to other worker nodes. 
The processing element that handles the global variable $b$ is sometimes called the central collector or the fusion center.

\section{Algorithm Description} \label{sec:mainAlg}
\subsection{Overall Algorithm}
We now describe the distributed trajectory optimization algorithm to solve the problem \eqref{eq:dist_cost}.
After applying initial conditions, our algorithm works iteratively based on the distributed ADMM with three major steps. First, the original problem \eqref{eq:dist_cost} is split into a separate optimization problem for each target and solved in parallel. Based on the optimization results, the common global variables are updated for consensus. Then, the Lagrange multipliers are updated to enforce the local variables to be closer to the common global variables.
The complete procedure is summarized in Algorithm \ref{alg:dto}.

To solve the augmented Lagrangian, the split optimization problem for each target $j$ is:
\begin{equation}
\begin{split}
\big[\bar{\textbf{P}}_j,\bar{\textbf{U}}_j \big] 
= &\argmin_{\textbf{P}_j,\textbf{U}_j} \sum_{t=0}^{T}  
%		tr\big(Q\Sigma_t^{(j)}(\textbf{p}_{t})\big)\\ 
\mathbb{U}^{(j)}(\textbf{p}_{j,t})
+ \dfrac{1}{M}\big(\textbf{u}_{j,t}^{\top} R \textbf{u}_{j,t} + \eta(\textbf{p}_{j,t})\big)\\
&\hspace{-0.8cm}+ \dfrac{\rho_{v}}{2} {\lVert{\textbf{p}_{j,t}-\textbf{P}_t^{C}+ \textbf{P}_{j,t}^{\lambda}}\rVert}^2
+ \dfrac{\rho_{u}}{2} {\lVert{\textbf{u}_{j,t}-\textbf{U}_t^{C}+ \textbf{U}_{j,t}^{\lambda}}\rVert}^2:\\
&\textbf{p}_{j,t+1} = \textbf{f}(\textbf{p}_{j,t},\textbf{u}_{j,t}),\enspace\textbf{b}_{t+1}^{(j)} = \textbf{g}(\textbf{b}_{t}^{(j)},\textbf{p}_{j,t}) +  \boldsymbol{\psi}_{j,t}^{(j)},\\
&\textbf{p}_{j,0} = \textbf{p}_{init} \quad and 
\quad \textbf{b}_{0}^{(j)} = \textbf{b}_{init}^{(j)}.
\end{split} \label{eq:main_traj_opt}
\end{equation}	
where $\textbf{p}_j$ and $\textbf{u}_j$ are the local variables for the states and control inputs of the mobile sensors, respectively.
%	\begin{equation}
%		\begin{split}
%			\big[\bar{\textbf{P}}_j,\bar{\textbf{U}}_j \big] = \argmin_{\textbf{P},\textbf{U}} \sum_{t=0}^{T}  tr\big(\Sigma_t^{(j)}(\textbf{p}_{t})\big) + \dfrac{1}{M}\textbf{u}_t^{\top} R_t \textbf{u}_t
%			+ \dfrac{\rho}{2} {\lVert{\textbf{p}_t-\textbf{P}_t^{C}+ \textbf{P}_{j,t}^{\lambda}}\rVert}^2
%			+ \dfrac{\rho}{2} {\lVert{\textbf{u}_t-\textbf{U}_t^{C}+ \textbf{U}_{j,t}^{\lambda}}\rVert}^2\\
%			s.t. \quad \textbf{p}_{t+1} = \textbf{f}(\textbf{p}_{t},\textbf{u}_{t}), \quad \textbf{p}_{0} = \textbf{p}_{init} \quad and \quad \Sigma_0^{(j)} = \Sigma_{init}^{(j)}
%		\end{split}
%	\end{equation}
%In trajectory optimization methods, this can be viewed as adding an additional regularizer that trajectories have to be recreatable by a policy?. 
This is a trajectory optimization problem with two additional quadratic cost terms 
%	that regulates the distance between common variables $\big[\textbf{P}_t^{C},\textbf{U}_t^{C}\big]$ and trajectory variables $\big[\textbf{p}_t,\textbf{u}_t\big]$
and can be solved with the existing trajectory optimization method described in Section \ref{subsec:iLQG}.
The split problems are independent and, then, able to be solved in parallel, as described in Section \ref{subsec:admm}.
In this work, the trajectory variables $\big[\textbf{P}_j,\textbf{U}_j \big]$ in \eqref{eq:main_traj_opt} are initialized by the method described in Section \ref{subsection:initial_guess}, and they are averaged to initialize the global common variables $\big[\textbf{P}^{C},\textbf{U}^{C}\big]$.
%	The optimization can be done in parallel.
%	This is a trajectory optimization problem with two additional quadratic cost terms and can be solved with existing trajectory optimization methods described in section \ref{subsec:iLQG}.

The updates for the common global variables are as follows:
\begin{equation}
\begin{split}
\textbf{P}^{C} = \frac{1}{M} \sum_{j=1}^{M}(\bar{\textbf{P}}_j + \textbf{P}_j^{\lambda}), \enspace
\textbf{U}^{C} = \frac{1}{M} \sum_{j=1}^{M}(\bar{\textbf{U}}_j + \textbf{U}_j^{\lambda}).
\end{split} \label{eq:consensus_phase}
\end{equation}
This is the mean consensus process that takes the average of the results for each trajectory optimization. 
{Then, the common global variables affect the update of the trajectory variables for each target $j$ (\ref{eq:main_traj_opt}), which causes the results of trajectory optimization for each target to converge to the same solution.}

The updates of the Lagrange multipliers for each target $j$ are as follows:
\begin{equation}
\begin{split}
\textbf{P}_j^{\lambda} = \textbf{P}_j^{\lambda} + \frac{\alpha_{v}}{\rho_{v}}(\bar{\textbf{P}}_j - \textbf{P}^{C}), \enspace
\textbf{U}_j^{\lambda} = \textbf{U}_j^{\lambda} + \frac{\alpha_{u}}{\rho_{u}}(\bar{\textbf{U}}_j - \textbf{U}^{C}).
\end{split} \label{eq:lambda}
\end{equation}
The important role of the multipliers is making high-level decisions, i.e., task-assignment.
They put more forces on local variables which are largely deviated from the global variables to be close by reducing its magnitude more.
By affecting the cost function in (\ref{eq:main_traj_opt}), the optimal trajectory solutions are guided into more informative regions. This process lets the task-assignment be automatically accomplished so that mobile sensors can further reduce the uncertainty of the targets.
%Thus, $\lambda$ can be seen as an alternative to high-level decision-making.
% because it serves to improve overall costs by reflecting the results of each of the other subproblems.}

\begin{algorithm}[t]
	\caption{Distributed Trajectory Optimization}\label{alg:DTO_alg} 
	\begin{algorithmic}[1]%\small
		\State Choose penalty weight $\rho$ and step-size $\alpha$
		\State Generate initial guess $\big[\bar{\textbf{P}}_j,\bar{\textbf{U}}_j \big]$ for each target $j$ 
		%		\State Initialize trajectory optimization solutions with $\big[\textbf{P}_j,\textbf{U}_j \big]$
		\While {not converged} 
		\State Update $\big[\bar{\textbf{P}}_j,\bar{\textbf{U}}_j \big]$ by solving each trajectory optimization problems (\ref{eq:main_traj_opt}) in parallel (see Section \ref{subsec:iLQG})
		\State Update $\big[\textbf{P}^{C},\textbf{U}^{C}\big]$ by averaging the results for each trajectory optimization
		\State Update $\big[\textbf{P}^{\lambda},\textbf{U}^{\lambda}\big]$ using (\ref{eq:lambda})
		\EndWhile
	\end{algorithmic} \label{alg:dto}
\end{algorithm}

\subsection{Trajectory Optimization} \label{subsec:iLQG}
We adopt belief space iterative Linear Quadratic Gaussian (belief space iLQG) \cite{van2012motion} to solve the trajectory optimization problems described in \eqref{eq:main_traj_opt}.		
Belief space iLQG extends the iLQG method\cite{todorov2005generalized} to Partially Observable Markov Decision Processes (POMDPs) using a Gaussian belief, instead of a fully observable state.
In the forward pass, belief space iLQG uses a standard EKF to compute the next time step belief. For a backward pass, belief space iLQG linearizes covariance in addition to quadratizing states and control inputs. Although the applications of this approach are focused on control problems, it is directly applicable to estimation problems (e.g., target tracking) due to the duality of control and estimation \cite{todorov2008general}

The belief space iLQG operates with the belief dynamics of the whole system with targets and sensors. That of the targets is given in (\ref{eq:belief_dyn}). By integrating it with the dynamics of mobile sensors, the belief dynamics of the entire system is represented by:
\begin{equation}
\boldsymbol{v}_{t+1} = \mathcal{F}(\boldsymbol{v}_{t},\textbf{u}_t) + \textbf{w}_t,\quad \textbf{w}_t \sim  \mathcal{N}(0,W),
\label{eq:entire_system}
\end{equation}
where 
\begin{align*}
\boldsymbol{v}_{t} = 
\begin{bmatrix}
\textbf{b}_{t}\\
\textbf{p}_{t}
\end{bmatrix}, 
\enspace
\mathcal{F}(\boldsymbol{v}_{t},\textbf{u}_t) = 
\begin{bmatrix}
\textbf{g}(\textbf{b}_{t},\textbf{p}_{t})\\
\textbf{f}(\textbf{p}_{t},\textbf{u}_{t})
\end{bmatrix},
\enspace
W = 
\begin{bmatrix}
\Psi & 0\\ 0 & 0
\end{bmatrix}.
\end{align*}

\subsubsection{Control Policy} 
Here, we explain the details of solving \eqref{eq:main_traj_opt} with the belief space iLQG.
By linearizing the dynamics around the nominal trajectory distribution, the approximate dynamics is expressed as:
\begin{equation}
\begin{split}
\boldsymbol{v}_{t+1} - \bar{\boldsymbol{v}}_{t+1} &\approx \mathcal{F}_{\boldsymbol{v},t}(\boldsymbol{v}_{t} - \bar{\boldsymbol{v}}_{t})
+ \mathcal{F}_{\textbf{u},t}(\textbf{u}_{t} - \bar{\textbf{u}}_{t}),\\
W_{(i)}(\boldsymbol{v}_{t},\textbf{u}_{t}) &\approx \textbf{e}_t^i + \mathcal{F}_{\boldsymbol{v},t}^{(i)}(\boldsymbol{v}_{t} - \bar{\boldsymbol{v}}_{t})
+ \mathcal{F}_{\textbf{u},t}^{(i)}(\textbf{u}_{t} - \bar{\textbf{u}}_{t}),
\end{split} \label{eq:linear_dyn}
\end{equation}
where
\begin{equation*}
\begin{split}
&\mathcal{F}_{\boldsymbol{v},t} = \frac{\partial \mathcal{F}}{\partial \boldsymbol{v}}(\bar{\boldsymbol{v}}_{t},\bar{\textbf{u}}_{t}), 
\enspace
\mathcal{F}_{\textbf{u},t} = \frac{\partial \mathcal{F}}{\partial \textbf{u}}(\bar{\boldsymbol{v}}_{t},\bar{\textbf{u}}_{t}),\\
&\enspace
\textbf{e}_t^i = W_{(i)}(\bar{\boldsymbol{v}}_{t},\bar{\textbf{u}}_{t}),
\enspace
\mathcal{F}_{\boldsymbol{v},t}^{i} = \frac{\partial W_{(i)}}{\partial \boldsymbol{v}}(\bar{\boldsymbol{v}}_{t},\bar{\textbf{u}}_{t}), \\
&\enspace
\mathcal{F}_{\textbf{u},t}^{i} = \frac{\partial W_{(i)}}{\partial \textbf{u}}(\bar{\boldsymbol{v}}_{t},\bar{\textbf{u}}_{t}). 				
\end{split}
\end{equation*}
$W_{(i)}(\boldsymbol{v}_{t},\textbf{u}_{t})$ is the $i$-th column of matrix $W(\boldsymbol{v}_{t},\textbf{u}_{t})$. Note that $W_{(i)}(\boldsymbol{v}_{t},\textbf{u}_{t})$ has $n$ columns, where $n$ is the dimension of the state.
We approximate the nonquadratic cost function in \eqref{eq:main_traj_opt} to quadratic one along the nominal belief and control trajectory	$(\bar{\boldsymbol{v}}, \bar{\textbf{u}})$. 
For the notional simplicity, the cost function \eqref{eq:main_traj_opt} is denoted by $\ell(\boldsymbol{v}_{t},\textbf{u}_{t})$,
\begin{equation}
\begin{split}
\ell(\boldsymbol{v}_{t},\textbf{u}_{t}) 
\approx&  \dfrac{1}{2} 
\begin{bmatrix}
\delta\boldsymbol{v}_{t} \\ \delta\textbf{u}_{t} 
\end{bmatrix}^{\top}
\begin{bmatrix}
\ell_{\boldsymbol{v}\boldsymbol{v},t} & \ell_{\boldsymbol{v}\textbf{u},t} \\ 	\ell_{\textbf{u}\boldsymbol{v},t} & \ell_{\textbf{u}\textbf{u},t}
\end{bmatrix}
\begin{bmatrix}
\delta\boldsymbol{v}_{t} \\ \delta\textbf{u}_{t} 
\end{bmatrix} \\
&+ 
\begin{bmatrix}
\delta\boldsymbol{v}_{t} \\ \delta\textbf{u}_{t}
\end{bmatrix}^{\top}
\begin{bmatrix}
\ell_{\boldsymbol{v},t} \\ \ell_{\textbf{u},t}
\end{bmatrix}
+ \ell_{0,t},
\end{split} \label{eq:quadratic_cost}
\end{equation}	
%with
%\begin{equation}
%\begin{split}
%&\ell(\boldsymbol{v}_{t},\textbf{u}_{t}) 
%= \sum_{t=0}^{T}  tr\big(Q\Sigma_t^{(j)}(\textbf{p}_{t})\big) 
%+ \dfrac{1}{M}\big(\textbf{u}_t^{\top} R \textbf{u}_t + \eta(\textbf{p}_{t})\big)\\
%&\quad + \dfrac{\rho}{2} {\lVert{\textbf{p}_t-\textbf{P}_t^{C}+ \textbf{P}_{j,t}^{\lambda}}\rVert}^2
%+ \dfrac{\rho}{2} {\lVert{\textbf{u}_t-\textbf{U}_t^{C}+ \textbf{U}_{j,t}^{\lambda}}\rVert}^2,
%\end{split} \label{eq:traj_cost}
%\end{equation}
where $\ell_{0,t} = \ell(\bar{\boldsymbol{v}}_{t},\bar{\textbf{u}}_{t})$. 
$\delta\boldsymbol{v}_{t} = \boldsymbol{v}_{t} - \bar{\boldsymbol{v}}_{t}$ and $\delta\textbf{u}_{t} = \textbf{u}_{t} - \bar{\textbf{u}}_{t}$ are the deviations from the nominal trajectory. The terms with subscripts denote Jacobian and Hessian matrices of their respective functions. 

\begin{figure*}[t] 
	\centering
	\subfigure[Modified dubins path]{
		\includegraphics[width=0.8\columnwidth]{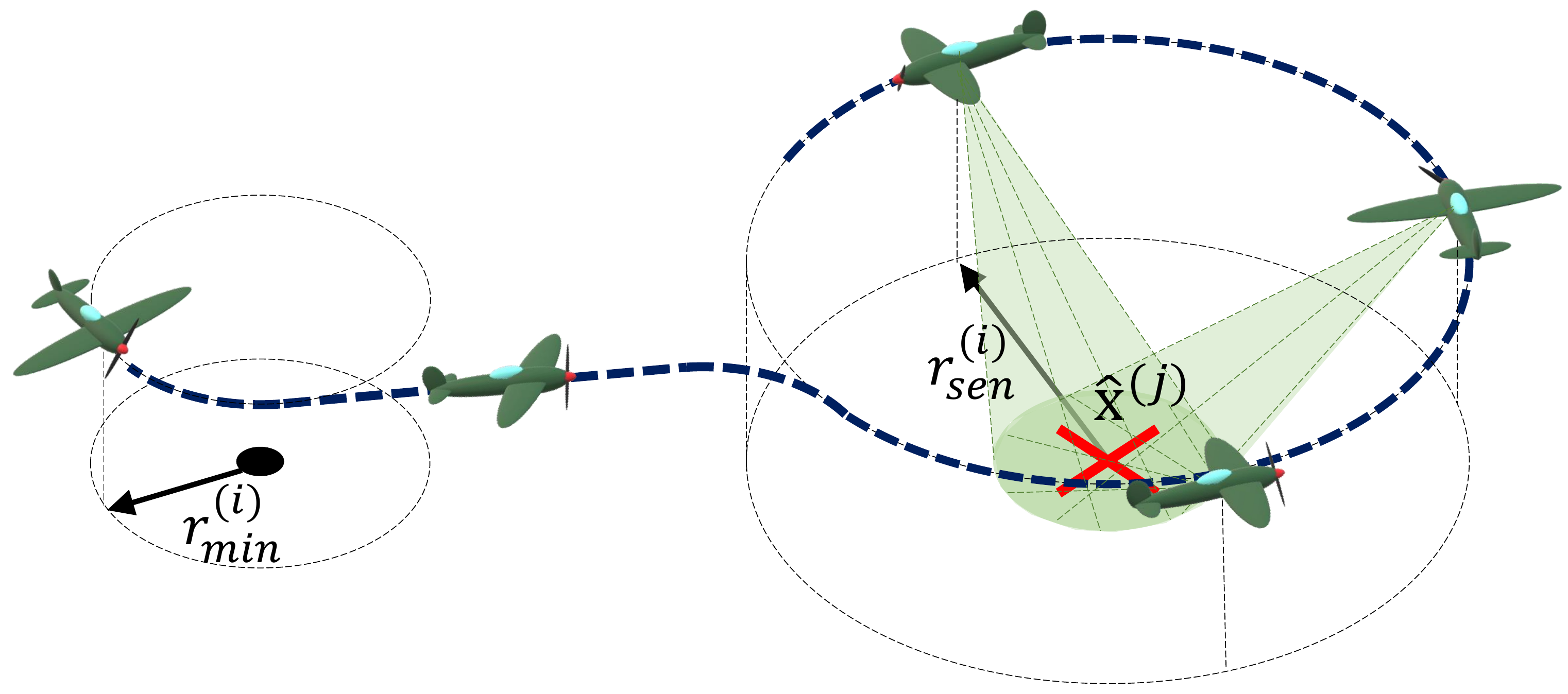}}
	\hspace*{1cm}%
	\subfigure[Path tracking scheme]{
		\includegraphics[width=0.8\columnwidth]{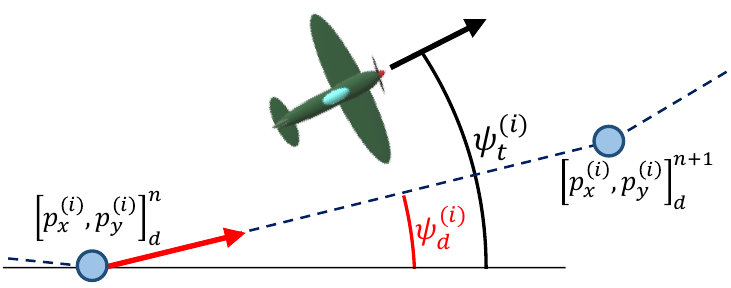}}	
	\caption{The path is generated by a modified Dubins path algorithm, which is used to provide waypoints for the generation of the initial guess.
		Control inputs are generated so that UAV follows the path connecting the waypoints through the PD-controller, and these are used as the initial guess of the local trajectory optimization method.}
	\label{fig:initial_guess_scheme}
\end{figure*} 

Given the linear dynamics \eqref{eq:linear_dyn} and quadratic cost \eqref{eq:quadratic_cost}, we obtain a quadratic approximation of the value function along a nominal trajectory $\bar{\boldsymbol{v}}_t$: 
\begin{equation}
\begin{split}
\boldsymbol{V}_{t}(\boldsymbol{v}_{t}) &\approx \dfrac{1}{2} \delta\boldsymbol{v}_{t}^{\top} \boldsymbol{V}_{\boldsymbol{v}\boldsymbol{v},t} \delta\boldsymbol{v}_{t} 
+ \delta\boldsymbol{v}_{t}^{\top} \boldsymbol{V}_{\boldsymbol{v},t} + \boldsymbol{V}_{0,t}.\\ 
%		\qquad \delta\boldsymbol{b}_{t} = \boldsymbol{b}_{t} - \boldsymbol{\bar{b}}_{t}
\end{split} \label{eq:ori_value_function}
\end{equation}		
Following the dynamic programming principle \cite{jacobson1970differential}, the Bellman equation for the value function $\boldsymbol{V}_{t}(\boldsymbol{v}_{t})$ and control policy $\pi_{t}(\boldsymbol{v}_{t})$ in discrete-time are specified as:
\begin{equation}
\begin{split}
\boldsymbol{V}_{t}(\boldsymbol{v}_{t})
&= \min_{\textbf{u}_t} \Big( \ell(\boldsymbol{v}_{t},\textbf{u}_{t}) + \mathbb{E} \big[ \boldsymbol{V}_{t+1}(\mathcal{F}(\boldsymbol{v}_{t},\textbf{u}_{t}) + \textbf{w}_t) \big] \Big) \\
&= \min_{\textbf{u}_t} \Big( \ell(\boldsymbol{v}_{t},\textbf{u}_{t}) + 
\dfrac{1}{2} \delta\boldsymbol{v}_{t+1}^{\top} \boldsymbol{V}_{\boldsymbol{v}\boldsymbol{v},t+1} \delta\boldsymbol{v}_{t+1} \\
&\enspace\quad\quad\quad+ \delta\boldsymbol{v}_{t+1}^{\top} \boldsymbol{V}_{\boldsymbol{v},t+1} + \boldsymbol{V}_{0,t+1} \\
&\enspace\quad\quad\quad+\dfrac{1}{2} tr 
\big[W(\boldsymbol{v}_{t},\textbf{u}_{t})^{\top} \boldsymbol{V}_{\boldsymbol{v}\boldsymbol{v},t+1} W(\boldsymbol{v}_{t},\textbf{u}_{t}) \big] \Big) \\
&\enspace= \min_{\textbf{u}_t} \boldsymbol{Q}(\boldsymbol{v}_{t}, \textbf{u}_t), \\
\pi_{t}(\boldsymbol{v}_{t})
&= \argmin_{\textbf{u}_t} \Big( \ell(\boldsymbol{v}_{t},\textbf{u}_{t}) + \mathbb{E} \big[ \boldsymbol{V}_{t+1}(\mathcal{F}(\boldsymbol{v}_{t},\textbf{u}_{t}) + \textbf{w}_t) \big] \Big),
\end{split} \label{eq:value_function}
\end{equation}	
%		where $\boldsymbol{V}_{t}(\boldsymbol{v}_{t})$ is the value function for the states $\boldsymbol{v}$ at time step $t$.
where 
\begin{equation*}
tr \big[W(\boldsymbol{v}_{t},\textbf{u}_{t})^{\top} \boldsymbol{V}_{\boldsymbol{v}\boldsymbol{v},t+1} W(\boldsymbol{v}_{t},\textbf{u}_{t}) \big] = \sum_{i=1}^{m} W_{(i)}(\boldsymbol{v}_{t},\textbf{u}_{t}).
\end{equation*}
By substituting equations \eqref{eq:linear_dyn} and (\ref{eq:quadratic_cost}) into (\ref{eq:value_function}), the Q-function is given by:
\begin{equation}
\begin{split}				
\boldsymbol{Q}_{t}(\bar{\boldsymbol{v}}_{t}+\delta\boldsymbol{v}_{t},\bar{\textbf{u}}_{t}
+\delta\textbf{u}_{t})
&= \dfrac{1}{2} 
\begin{bmatrix}
\delta\boldsymbol{v}_{t} \\ \delta\textbf{u}_{t} 
\end{bmatrix}^{\top}
\begin{bmatrix}
Q_{\boldsymbol{v}\boldsymbol{v},t} & Q_{\boldsymbol{v}\textbf{u},t} \\ 	Q_{\textbf{u}\boldsymbol{v},t} & Q_{\textbf{u}\textbf{u},t}
\end{bmatrix}
\begin{bmatrix}
\delta\boldsymbol{v}_{t} \\ \delta\textbf{u}_{t} 
\end{bmatrix}\\
&\quad+ 
\begin{bmatrix}
\delta\boldsymbol{v}_{t} \\ \delta\textbf{u}_{t}
\end{bmatrix}^{\top}
\begin{bmatrix}
Q_{\boldsymbol{v},t} \\ Q_{\textbf{u},t}
\end{bmatrix}
+ Q_{0,t},						
\end{split} \label{eq:q_fuction}
\end{equation}		
where
\begin{equation} 
\begin{split}
&\boldsymbol{Q}_{\boldsymbol{v}\boldsymbol{v},t} = \ell_{\boldsymbol{v}\boldsymbol{v},t} + \mathcal{F}_{\boldsymbol{v},t}^{\top} \boldsymbol{V}_{\boldsymbol{v}\boldsymbol{v},t+1}\mathcal{F}_{\boldsymbol{v},t}
+ \sum_{i=1}^{m} \mathcal{F}_{\boldsymbol{v},t}^{i{\top}} \boldsymbol{V}_{\boldsymbol{v}\boldsymbol{v},t+1} \mathcal{F}_{\boldsymbol{v},t}^{i},\\
&\boldsymbol{Q}_{\boldsymbol{v},t} = \ell_{\boldsymbol{v},t} + \mathcal{F}_{\boldsymbol{v},t}^{\top} \boldsymbol{V}_{\boldsymbol{v},t+1} 
+ \sum_{i=1}^{m} \mathcal{F}_{\boldsymbol{v},t}^{i{\top}} \boldsymbol{V}_{\boldsymbol{v}\boldsymbol{v},t+1} \textbf{e}_{t}^{i},\\
&\boldsymbol{Q}_{\textbf{u}\textbf{u},t} = \ell_{\textbf{u}\textbf{u},t} + \mathcal{F}_{\textbf{u},t}^{\top} \boldsymbol{V}_{\boldsymbol{v}\boldsymbol{v},t+1}\mathcal{F}_{\textbf{u},t}
+ \sum_{i=1}^{m} \mathcal{F}_{\textbf{u},t}^{i{\top}} \boldsymbol{V}_{\boldsymbol{v}\boldsymbol{v},t+1} \mathcal{F}_{\textbf{u},t}^{i},\\
&\boldsymbol{Q}_{\textbf{u},t} = \ell_{\textbf{u},t} + \mathcal{F}_{\textbf{u},t}^{\top} \boldsymbol{V}_{\boldsymbol{v},t+1}
+ \sum_{i=1}^{m} \mathcal{F}_{\textbf{u},t}^{i{\top}} \boldsymbol{V}_{\boldsymbol{v}\boldsymbol{v},t+1} \textbf{e}_{t}^{i},\\
&\boldsymbol{Q}_{\textbf{u}\boldsymbol{v},t} = \ell_{\textbf{u}\boldsymbol{v},t} + \mathcal{F}_{\textbf{u},t}^{\top} \boldsymbol{V}_{\boldsymbol{v}\boldsymbol{v},t+1}\mathcal{F}_{\boldsymbol{v},t}
+ \sum_{i=1}^{m} \mathcal{F}_{\textbf{u},t}^{i{\top}} \boldsymbol{V}_{\boldsymbol{v}\boldsymbol{v},t+1} \mathcal{F}_{\boldsymbol{v},t}^{i},\\	
&\boldsymbol{Q}_{0,t} = \ell_{0,t} + \boldsymbol{V}_{0,t+1} + \sum_{i=1}^{n} \textbf{e}_t^{i{\top}} \boldsymbol{V}_{\boldsymbol{v}\boldsymbol{v},t+1} \textbf{e}_t^{i}.	
\end{split} \label{eq:q_fuction2}
\end{equation}

In order to find the optimal control policy, we compute the local variations in the control $\delta\hat{\textbf{u}}$ that minimize the quadratic approximation of the $Q$-function in (\ref{eq:value_function}):
\begin{equation}
\begin{split}
\delta\hat{\textbf{u}}_{t} &= \argmax \big[ \boldsymbol{Q}_{t}(\bar{\boldsymbol{v}}_{t}+\delta\boldsymbol{v}_{t},\bar{\textbf{u}}_{t}+\delta\textbf{u}_{t}) \big]\\
&= -\boldsymbol{Q}_{\textbf{u}\textbf{u},t}^{-1}\boldsymbol{Q}_{\textbf{u},t} -\boldsymbol{Q}_{\textbf{u}\textbf{u},t}^{-1}\boldsymbol{Q}_{\textbf{u}\boldsymbol{v},t}\delta\boldsymbol{v}_{t}.
\end{split}\label{eq:optimal_policy}
\end{equation} 
%		This recurrence allows us to compute the optimal control law as 
The optimal control can be found as $\hat{\textbf{u}}_{t} = \bar{\textbf{u}}_{t} + \delta\hat{\textbf{u}}_{t}$.
Substituting (\ref{eq:optimal_policy}) into (\ref{eq:value_function}) gives the value function $\boldsymbol{V}_{t}(\boldsymbol{v}_{t})$ as a function of only $\boldsymbol{v}_{t}$ in the form of (\ref{eq:ori_value_function}):
\begin{equation}
\begin{split}
\boldsymbol{V}_{\boldsymbol{v}\boldsymbol{v},t} &= \boldsymbol{Q}_{\boldsymbol{v}\boldsymbol{v},t} 
-\boldsymbol{Q}_{\textbf{u}\boldsymbol{v},t}^{\top}\boldsymbol{Q}_{\textbf{u}\textbf{u},t}^{-1} \boldsymbol{Q}_{\textbf{u}\boldsymbol{v},t},\\
\boldsymbol{V}_{\boldsymbol{v},t} &= \boldsymbol{Q}_{\boldsymbol{v},t} 
-\boldsymbol{Q}_{\textbf{u}\boldsymbol{v},t}^{\top}\boldsymbol{Q}_{\textbf{u}\textbf{u},t}^{-1} \boldsymbol{Q}_{\textbf{u},t},\\
\boldsymbol{V}_{0,t} &= \boldsymbol{Q}_{0,t} 
-\boldsymbol{Q}_{\textbf{u},t}^{\top}\boldsymbol{Q}_{\textbf{u}\textbf{u},t}^{-1} \boldsymbol{Q}_{\textbf{u},t}.			
\end{split}
\end{equation}
%		We can express them with the following recurrences,
This recursion then continues by computing the control policy for time step $t-1$.

\subsubsection{Initial Guess Generation} \label{subsection:initial_guess}

%In the local trajectory optimization, the initial guess has a great effect on the convergence speed as well as on the quality of the solution. 
%Thus, it is very important to generate appropriate initial guesses for the purpose of the problem.
%To do this, we use a Dubins path\cite{dubins1957curves} that can mimic the behavior of a fixed-wing UAV and slightly modify it to fit the target tracking problem. 
For fast convergence to a better solution, good initial guesses are still required in the subproblems \eqref{eq:main_traj_opt}, in which it is much easier to find a good one than the original problem \eqref{eq:original_cost}.
For the initialization, we derive an initial guess generator for a case of fixed-wing UAV mounting a sensor with perpendicular boresight. The generator is based on Dubins paths\cite{dubins1957curves} that can mimic the behavior of a fixed-wing UAV.

We first generate a continuous path. Let $ r_{min}^{(i)}$, $r_{sen}^{(i)}$ and $\hat{x}^{(j)}$ be the minimum turning radius of the $i$-th UAV, the distance between the nadir point and the center point of the $i$-th UAV's sensor footprint and the expected position of the target $j$, respectively. 
As shown in Fig. \ref{fig:initial_guess_scheme}(a), a path is created to arrive as soon as possible at any point of a circle with radius $r_{sen}^{(i)}$ centered on $\hat{x}^{(j)}$ with the sensor boresight toward the center. Then, the remaining path is set for the sensor to rotate along the circle.
Here, $r_{sen}^{(i)}$ should be set to an appropriate value reflecting the movement of the target. The expected position $\hat{x}^{(j)}$ of the target $j$ can be predicted using the current estimates and the dynamic model of the target. 

Then, the generated path is discretized at specific intervals and used as waypoints to generate a sequence of control inputs.
For the sequence of desired waypoints $([p_x^{(i)},p_y^{(i)}]_d^1,\cdots,[p_x^{(i)},p_y^{(i)}]_d^n,[p_x^{(i)},p_y^{(i)}]_d^{n+1},\cdots)$,
we construct a controller that reduces the angle between the line connecting the two adjacent waypoints and the heading direction of the UAV. A damping effect is also added to the controller for the stability. The geometry of this path tracking problem is depicted in Fig. \ref{fig:initial_guess_scheme}(b). In our simulation, the control input of the PD-controller can be computed with \eqref{eq:uav_dyn} as follows: 
\begin{equation}
u_{t+1}^{(i)} = -K_p(\psi^{(i)}_t-\psi^{(i)}_{d}) - K_d \phi^{(i)}_t,
\end{equation}
where $K_p$ and $K_d$ are the gains of the controller.

%\XX{  	
%That algorithm determines the penalty weight $\rho$ defined by the infinite, constant, and zero in the consensus process based on the results of each subproblems.
%Our approach differs in that it can be considered to use only constant and zero in the message-passing algorithm, but excludes targets whose             uncertainty can not be reduced beyond a specified $\epsilon$ value during the iteration of the distributed optimization.
%Specifically, the three weight message-passing algorithm maintains graphs during the iteration of ADMM, but our approach transforms the graphs through cutting 
%as shown in Fig. \ref{fig:graph}(b).}

\subsection{Modified Receding Horizon Control for Real-Time Operation} \label{subsec:planning}

For real-time operation, the planning should be done while the UAVs are executing the previous plan. Otherwise, they stop and wait for the next plan for every planning cycle. Also, the established plan could be corrected by reflecting newly obtained information during its execution.
For these purposes, we propose a modified receding-horizon control (RHC) scheme with introducing a new concept of control horizon. 
RHC is a form of a feedback control system which executes only the forepart of a plan and start over the planning for a shifted time horizon with updated information \cite{kuwata2004three}.
Extending this concept, our algorithm plans for future $T_p$ time steps while executing the previous plan for $T_c$ time steps with predefined control horizon $T_c$ and planning horizon $T_p$ $(T_c\leq T_p)$.
Note that $T_c$ is determined by the available computational resources and the communication delay, and $T_p$ should be large enough so that the trajectories of the mobile sensors can cover the overall domain of interest.
%It executes only the forepart of a plan and start over the planning for a shifted time horizon with updated information. This increases the quality of the plan. In out problem, the planning process takes time due to its complexity and UAVs' motion are continuous. For this 

%In a multi-target tracking problem, prior information about the targets is limited and the computation time for trajectory planning can become impractically long as the complexity of the problem increases. Thus, a receding-horizon control scheme (RHC) can be appropriate. RHC, also known as model predictive control (MPC), is a form of feedback (i.e., closed-loop) control system. 
%With RHC, an optimization problem is solved at specific time intervals to determine a plan over a fixed time horizon and then the forepart of this plan is applied.
%The planning process is repeated by solving a new optimization problem, with the time horizon shifted a specific step forward.
%Optimization is performed based on available measurements and data at each step.
%Thus, the control policy involves feedback.
%The feedback control scheme enables real-time measurements to be used to determine the control input, and it can compensate for a deviation between the predicted and actual output that can be caused by system-model mismatches and disturbances.

The modified RHC is described in Algorithm \ref{alg:RA}. Here, we explain a single cycle starting at $t=1$.
%In situations where a non-linear tracking filter is utilized for the estimation of the target information, we approximate a target distribution into a Gaussian distribution for planning purposes (line 6).  
%One important point during the real-time implementation of the algorithm is that the future plan should be pre-calculated, which requires future information of targets.
Pre-calculating the future plan for $t=T_c$ to $T_c+T_p$ requires the target belief estimation at $t=T_c$. We use the maximum likelihood assumption and estimate the future belief without actual measurements (line 6).
A new plan $\big[\bar{\textbf{P}},\bar{\textbf{U}} \big]_{new}$ for $t=T_c$ to $T_c+T_p$ is created using the proposed distributed trajectory optimization algorithm (line 7). 
This planning is done in parallel to executing the previous plan.
%Here, we use the estimated future information of targets under the maximum likelihood assumption simultaneously with the previous plan at the control horizon $T_c$.
While the algorithm plans for $T_p$ future time steps, only the fore $T_c$ steps are actually executed (line 9).
The process is then repeated for the next time window. 

\begin{algorithm}[t]
	\caption{Modified Receding-Horizon Control}\label{alg:RA} 
	\begin{algorithmic}[1] %\small
		\State Set control horizon $T_c$ and planning horizon $T_p$  		
		\State Input $\textbf{p}_{1}$, $\textbf{b}_{1}$ and initial plan $\big[\bar{\textbf{P}},\bar{\textbf{U}} \big]$
		\While {operation time is remaining} 
		\For {$t\leftarrow1 \enspace \textbf{to} \enspace T_c -1$}
		\If {$t=1$} (in parallel)
		%					\State$\hat{\textbf{b}} = \textsc{gaussian approximation}(\textbf{b}_{t+1})$
		\State$\hat{\textbf{b}}_{T_c}, \textbf{p}_{T_c} = \textsc{propagate}(\textbf{b}_{1},\textbf{p}_{1},\big[\bar{\textbf{P}},\bar{\textbf{U}} \big])$
		\State$\big[\bar{\textbf{P}},\bar{\textbf{U}} \big]_{new} = \textsc{\normalsize create plan}(\hat{\textbf{b}}_{T_c},\textbf{p}_{T_c},T_p)$
		%					\State$\big[\bar{\textbf{P}},\bar{\textbf{U}} \big]_{new}$ % \linebreak
		%					\Statex $~~~~~~~~~~~~~~~~~~= \textsc{create plan}(\hat{\textbf{b}}_{T_c},\big[\bar{\textbf{P}}_{T_c},\bar{\textbf{U}}_{T_c} \big],T_p)$ 
		\Statex \Comment {Algorithm \ref{alg:DTO_alg}.}
		\EndIf
		\State $\textbf{b}_{t+1}=\textsc{execute plan}(\textbf{b}_{t},\big[\bar{\textbf{P}},\bar{\textbf{U}} \big],t)$
		\EndFor
		\State $\textbf{p}_{1},\textbf{b}_{1} = \textbf{p}_{T_c},\textbf{b}_{T_c}$
		\State $\big[\bar{\textbf{P}},\bar{\textbf{U}} \big] = \big[\bar{\textbf{P}},\bar{\textbf{U}} \big]_{new}$
		\EndWhile
	\end{algorithmic}
\end{algorithm}

%\subsection{Comparison with Exisiting Works} \label{subsec:planning}
%\XX{ As noted earlier, the recent approaches of [?] and concurrent [?] are related to ours.
%	In \cite{oh2013coordinated,farmani2017scalable,luders2011information}, methods for clustering targets have been proposed. The cost function of the optimization problem for clustering is defined using the distance between the targets and the sensor platforms and/or the uncertainty of the targets. Luders et al. \cite{luders2011information} combined a sampling-based path planning algorithm with a task allocation algorithm, the Consensus-Based Bundle Algorithm (CBBA\cite{choi2009consensus}).
%	In order to make it easier to deal with the task-assignment problem, 
%	%	this algorithm has defined the cost function by neglecting the correlation between measurements of sensors and then performed the task-assignment. 
%	they defined the cost function by neglecting the correlation between measurements of sensors before performing the task-assignment. 
%	Based on the results, they used a sampling-based path planning algorithm called the Information-rich Rapidly-exploring Random Tree (IRRT) algorithm~\cite{levine2010information} for trajectory planning in the presence of obstacles.}

\section{Complexity Reduction} \label{sec:complexity}	
\subsection{Complexity Analysis} 
The computational complexity of each iteration of the distributed ADMM is dominated by solving the subproblems defined in \eqref{eq:main_traj_opt} because computing the global common variable and Lagrange multipliers requires very simple computations.
%		In our application of ADMM, computational complexity is dominated by solving the subproblems defined in eq. \eqref{eq:main_traj_opt} because computing the global common variable and lagrange multipliers requires very simple computations.
The subproblems are solved by the belief space iLQG method described in Section \ref{subsec:iLQG} , and its computational complexity in control problems is already well analyzed in \cite{van2012motion}.
The bottleneck of running time in our problem lies in the calculation of the matrix $\boldsymbol{Q}_{\boldsymbol{v}\boldsymbol{v},t}$ in \eqref{eq:q_fuction}.

We first consider the case of a single target. Let $m$ and $n$ be the state dimensions of the target and the mobile sensors, respectively.	
As the target belief contains the covariance matrix of the state, the dimension of the belief is $O[m^2]$.
Accordingly, the belief dimension of the entire system is $O[m^2+n]$, and
%		If $m^2$ is assumed to be greater than $n$, 
computing the product $\mathcal{F}_{\boldsymbol{v},t}^{\top} \boldsymbol{V}_{\boldsymbol{v}\boldsymbol{v},t+1}\mathcal{F}_{\boldsymbol{v},t}$ in \eqref{eq:q_fuction2}
takes $O[(m^2+n)^3]$ time. Evaluating $\ell_{\boldsymbol{v}\boldsymbol{v},t}$ using numerical differentiation (central differences) can be done in $O[(m^2+n)^3]$ time.
The remaining elements do not form bottlenecks during computation as discussed in \cite{van2012motion}.		
Then, since one cycle of value iteration takes $T$ steps, its complexity is $O[T(m^2+n)^3]$. 
The number of cycles cannot be expressed in terms of dimensional notation, but convergence can be expected after 10-100 cycles in practice.
Furthermore, it is known that belief space iLQG converges with a second-order rate to a local optimum \cite{van2012motion}. 
%		\RE{For the sake of easy comparison, we assume that $m^2$ is greater than $n$.} 

For multiple targets, our algorithm operates distributedly. The complexity is, then, multiplied by the number of iterations of distributed ADMM if the computation is fully parallelized. ADMM is known to require a few tens of iterations to converge with modest accuracy for many applications \cite{boyd2011distributed,liu2014optimal,masazade2012sparsity}. Note that when the original problem \eqref{eq:original_cost} is solved non-distributedly, the complexity is multiplied by $M^2$, instead, regarding the independence among the targets. This is more complex than our distributed approach when $M$ is large, and it even requires high-level decision-making as mentioned in Section~\ref{subsec:challenge}. There have been many researches on fast convergence for general distributed ADMM\cite{bento2013message,derbinsky2013improved,song2016fast}. In this paper, we propose an edge-cutting method which is tailored to our algorithm and effectively shortens its running time.

\begin{figure}[t]
	\centering
	\subfigure[]{
		\includegraphics[width=4.cm]{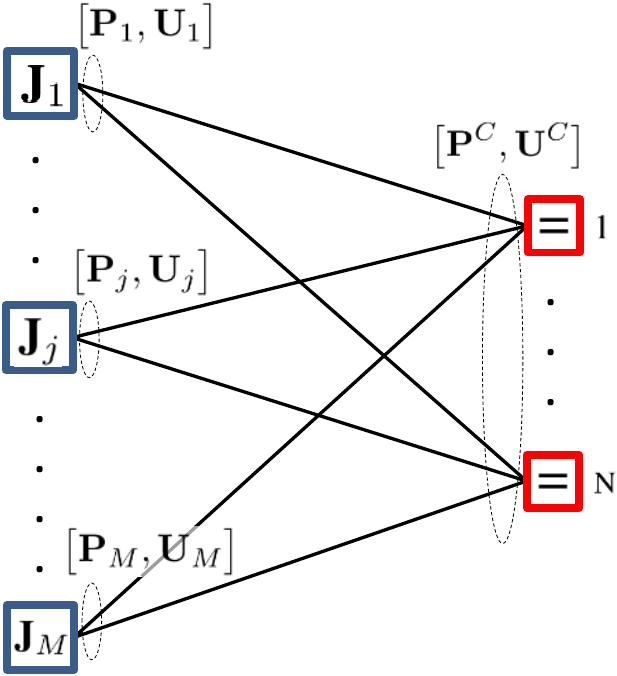}}
	\hspace*{.2cm}%
	\subfigure[]{
		\includegraphics[width=4.cm]{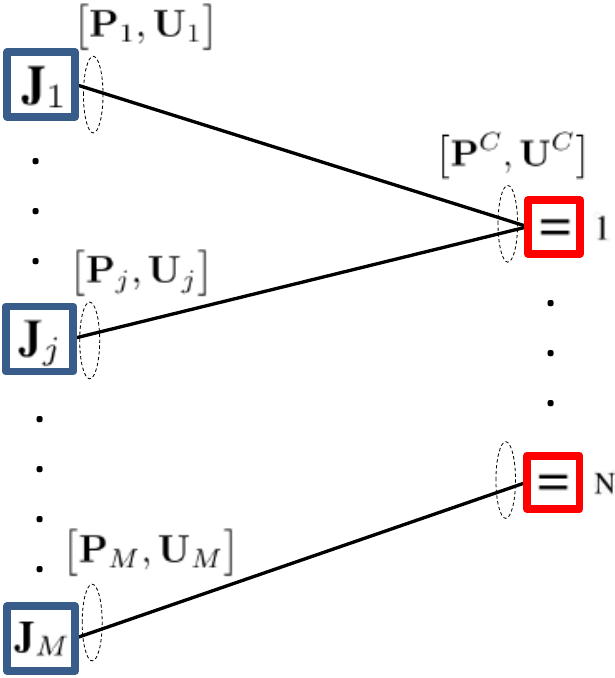}}			
	\caption{(a) Fully connected bipartite graph, and (b) bipartite graph after applying the edge-cutting method}
	\label{fig:graph}
\end{figure}

\subsection{Edge-cutting Method} \label{subsec:Heuristic_rule}

The computational load of our algorithm can be effectively reduced when we consider the characteristics of our problem. As pointed out earlier, our distributed scheme automatically makes high-level decisions (i.e., task-assignment). A mobile sensor may choose a path specialized for certain targets during the consensus and disregards the others. In this case, the sensor cannot contribute to tracking the disregarded targets, and, thus, it is a bad investment of computational resources to solve the trajectory optimization of the sensor for those targets. Based on this observation, we propose a heuristic method to detect this inefficacy and halt the unprofitable investment.

% 	When a mobile sensor tracks multiple targets within a given planning time, it might be better to track some of them. Also, it may be more efficient for mobile sensors to track targets separately when multiple sensors track multiple targets. Empirically, these output characteristics can be obtained by the proposed algorithm.
% 	The proposed algorithm, however, continues to solve the trajectory optimization problem between a mobile sensor and a target whose uncertainty is not reduced by the mobile sensor during the iteration of the distributed optimization, which slows down the computation.
% 	To tackle this problem, we propose an edge-cutting method which cuts an edge and does not solve unnecessary problems (Fig. \ref{fig:graph}).
The edge-cutting method detects unavailing sensors for each target for every consensus step in ADMM.
For given $\epsilon>0$, the method excludes sensor $i$ in the tracking task of target $j$ if
\begin{equation} 
\begin{split}
&\sum_{t=0}^{T}  \Expec_{\textbf{z}_t^{(j)}} \Big[ tr\big(\Sigma_t^{(j)}(\textbf{p}_{t}^{-(i)})\big) - tr\big(\Sigma_t^{(j)}(\textbf{p}_{t})\big) \Big]<\epsilon\\
\end{split}, 
\label{eq:evaluate_traj}
\end{equation}		
where $\mathbb{E}_{\textbf{z}_t^{(j)}} \Big[ tr\big(\Sigma_t^{(j)}(\textbf{p}_t^{-(i)})\big)\Big]$ denotes the uncertainty of the target $j$ when the sensor $i$ is excluded.
%	$\textbf{p}_t^{-(i)}$ is given, $\textbf{p}_t^{-(i)}$ denotes the states of the mobile sensors except mobile sensor $i$ 
%	at time $t$. $tr\big(\Sigma_t^{(j)}(\textbf{p}_{t})\big)$ is the amount of uncertainty about a target $j$ when the states of all mobile sensors $\textbf{p}_t$ are given. 
% 	If the evaluated value is less than the specified value $\epsilon$, the corresponding edge is cut.
The $\epsilon$ is chosen to be a threshold for judging whether the sensor $i$ reduces the uncertainty of the target $j$ enough or not. The uncertainty is evaluated for $\big[\textbf{P}^{C},\textbf{U}^{C}\big]$ obtained in the consensus phase \eqref{eq:consensus_phase} except for the sensor $i$.
This pruned architecture is reset to the original one for new time windows. When $\epsilon$ is properly chosen (not too large), the method has empirically shown similar target estimation results as not applying it in most cases with significantly reduced computation time. The method is quantitatively evaluated in Section~\ref{sec:sim_edge}.
% 	Evaluating \eqref{evaluate traj} requires an estimate of the target states.
% 	\RE{As in Section \ref{subsec:estimation}}, we apply the maximum-likelihood observations assumption to obtain an estimate of the target state in the absence of measurements.
% 	It is assumed that all trajectories except for the trajectory of the $i$-th mobile sensor use the trajectory obtained in the consensus phase given by \eqref{eq:consensus_phase}.
% 	This method may sometimes generate a suboptimal solution, but it empirically appears to yield a result identical the best solution in most situations. 
% 	Problems that arise when using the edge-cutting method can be solved by a receding-horizon control scheme (Section \ref{subsec:planning}), and if  $\epsilon$ is well adjusted, a reasonable result can be obtained.
% 	We note that the proposed method has the purpose of shortening the computation time of the problem, and if $\epsilon$ is not set to be too large, it provides the same result as the algorithm without the edge-cutting method.

Our edge-cutting method has similar, but different, concept with the three weight message-passing algorithm (TWA) in \cite{bento2013message}, which solves multi-agent trajectory optimization for energy minimization and collision-avoidance. As shown in Fig.~\ref{fig:graph}, our algorithm can be also expressed with a bipartite graph as in \cite{bento2013message} based on the structure of ADMM. Each blue node on the left side represents the trajectory optimizer solving \eqref{eq:main_traj_opt} for each target, and each red node on the right side is the consensus node solving \eqref{eq:consensus_phase} for each sensor. For the consensus node, TWA employs weighted averages and excludes an edge from the consensus process by applying a zero weight without any change in the graph. On the other hand, our method prunes an edge of the graph when \eqref{eq:evaluate_traj} is satisfied. This effectively reduces the complexities of the EKF in Section~\ref{subsec:estimation} and the iLQG in Section~\ref{subsec:iLQG} by reducing the dimensions of the sensor states and the measurement vector.
Note that our algorithm allows pruning an edge since the trajectory optimization for a target always produces feasible solutions when solved for subsets of sensors. This is not the case for the optimizers for collision-avoidance in \cite{bento2013message}.

\section{SIMULATIONS} \label{sec:simulation}

In this section, we numerically investigate the computational properties and the applicability of the proposed algorithms.
To ascertain the effectiveness of the distributed trajectory optimization algorithm, we compare it with existing approaches; non-myopic trajectory planning methods with heuristic task-assignments as well as the myopic one.
We also evaluate the modified RHC scheme and the edge cutting algorithm which are proposed for effective real-time operations.

\begin{figure}[t]
	\centering
	\includegraphics[width=8cm]{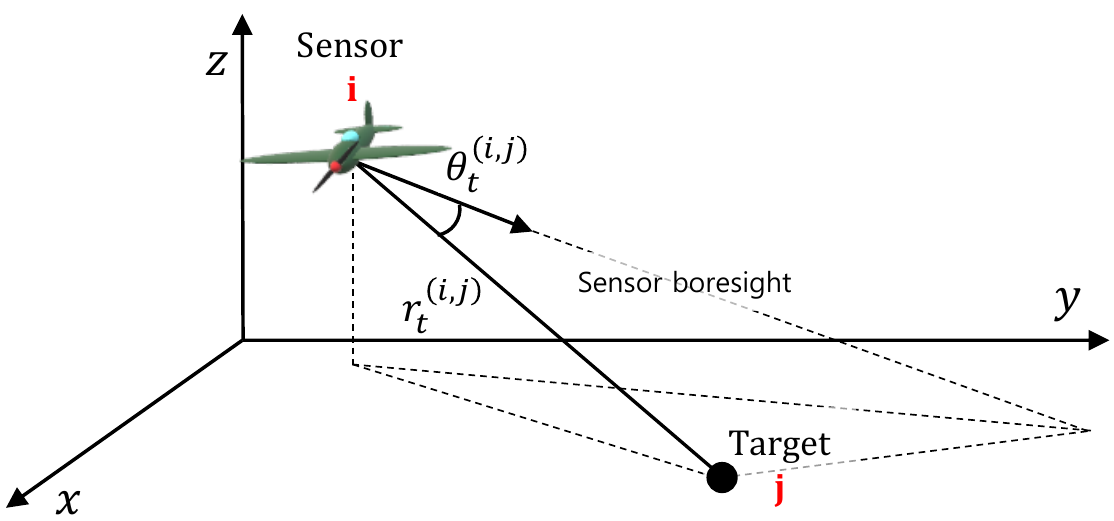}
	\caption{Sensor model for target tracking}
	\label{fig: sensor_geo}
\end{figure}

\subsection{Simulation Model}

\begin{figure*}[t]\centering{ 
		\subfigure[Initial guess]{
			\includegraphics[width=.5\columnwidth]{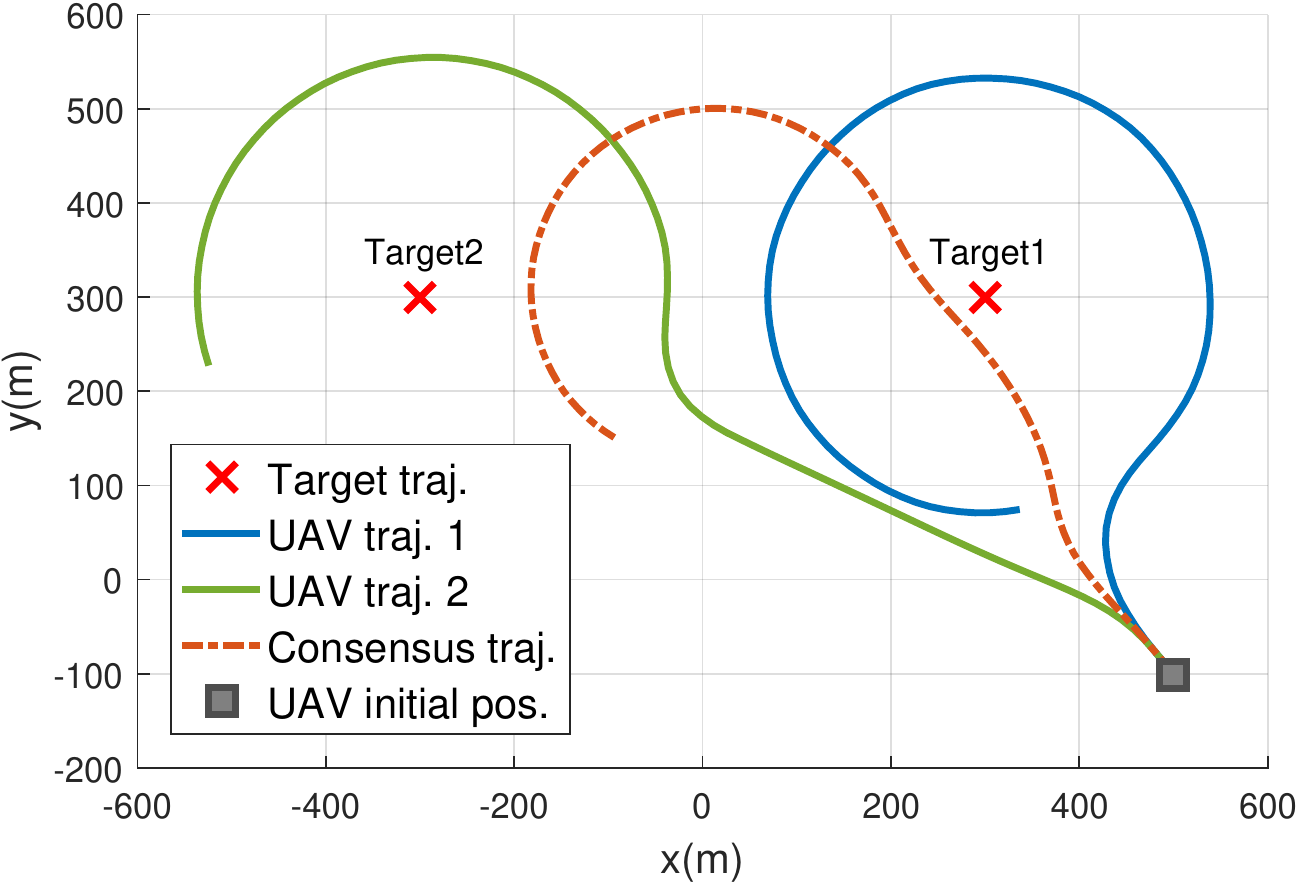}}
		\hspace*{-0.3cm}
		\subfigure[Number of iterations = 2]{
			\includegraphics[width=.5\columnwidth]{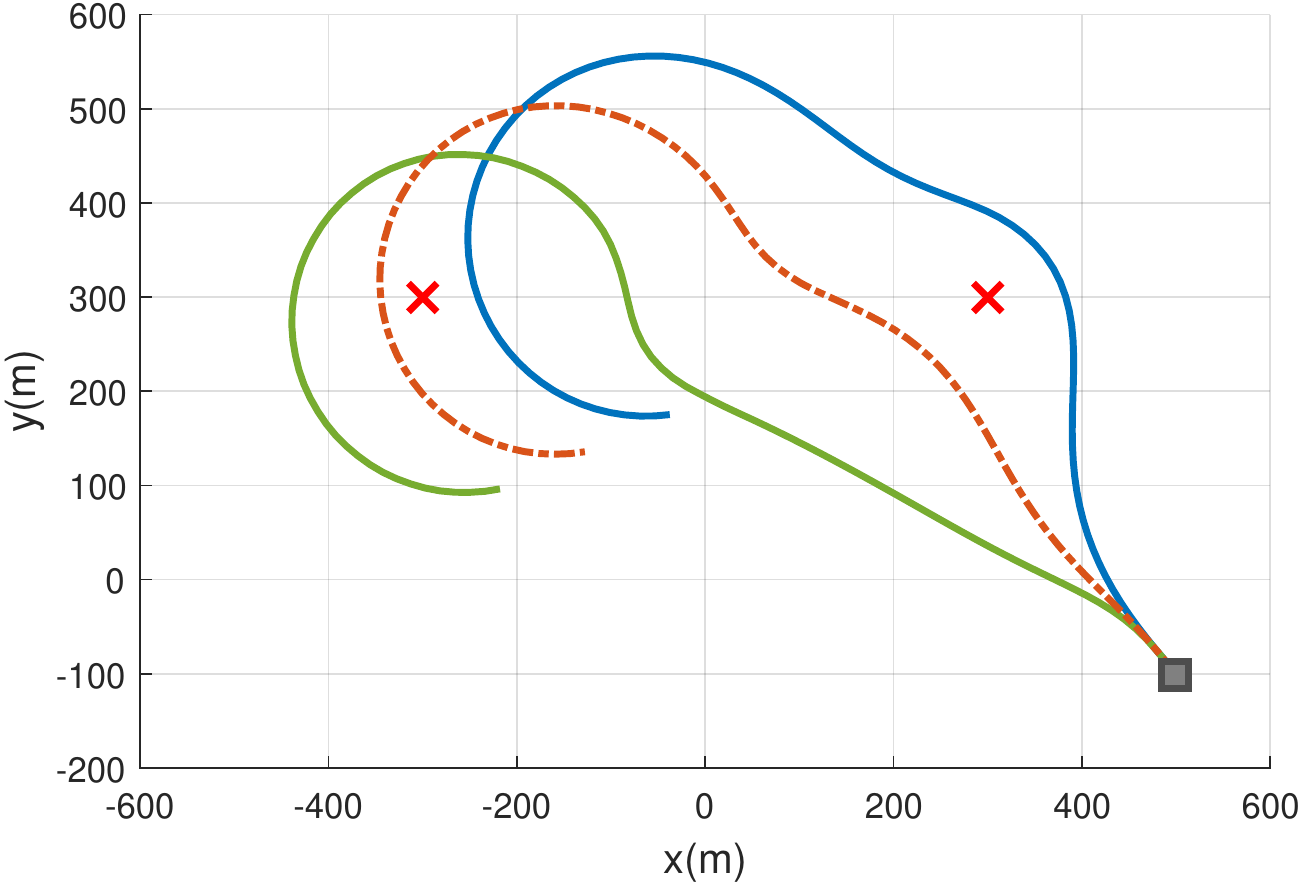}}
		\hspace*{-0.3cm}	
		\subfigure[Number of iterations = 5]{
			\includegraphics[width=.5\columnwidth]{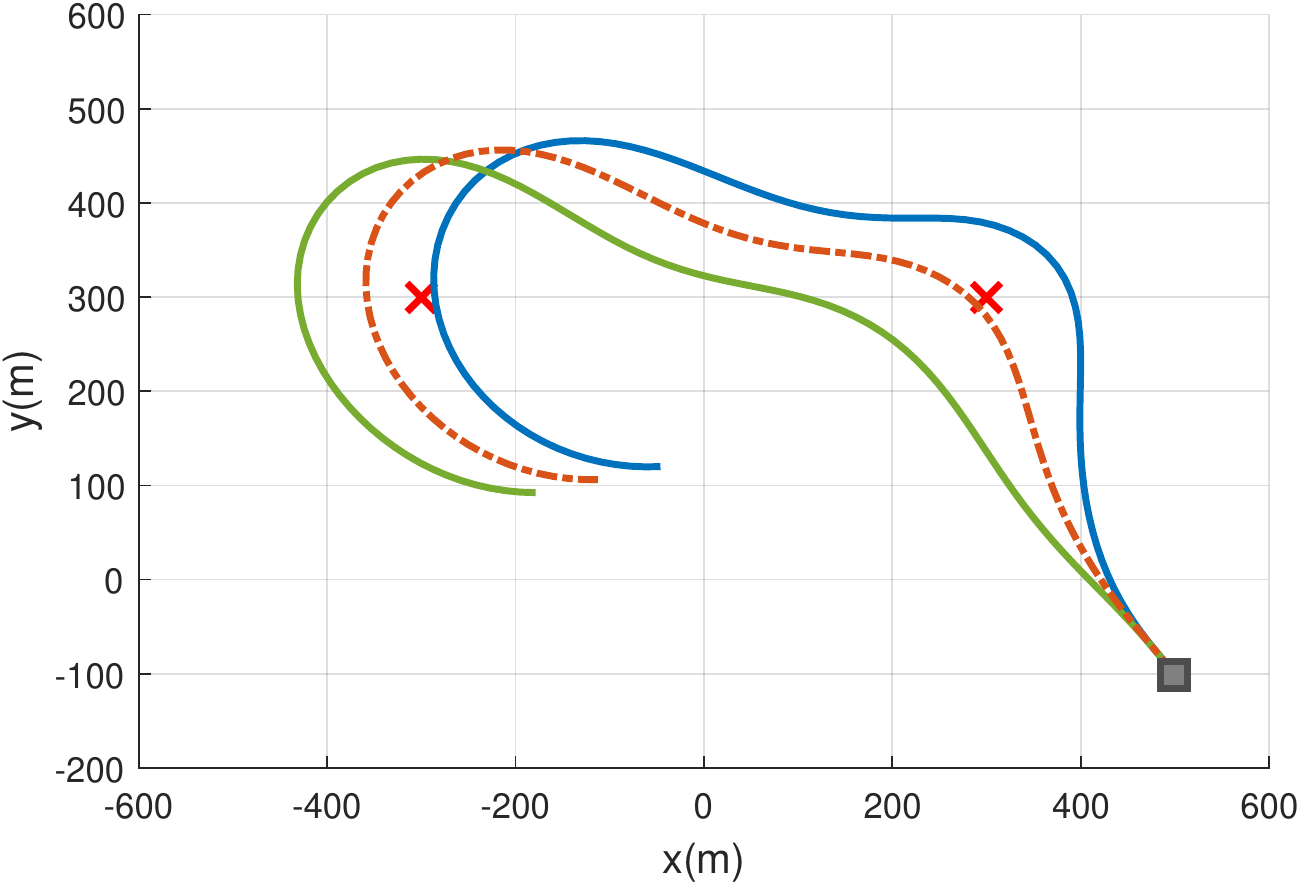}}
		\hspace*{-0.3cm}
		\subfigure[Number of iterations = 25]{
			\includegraphics[width=.5\columnwidth]{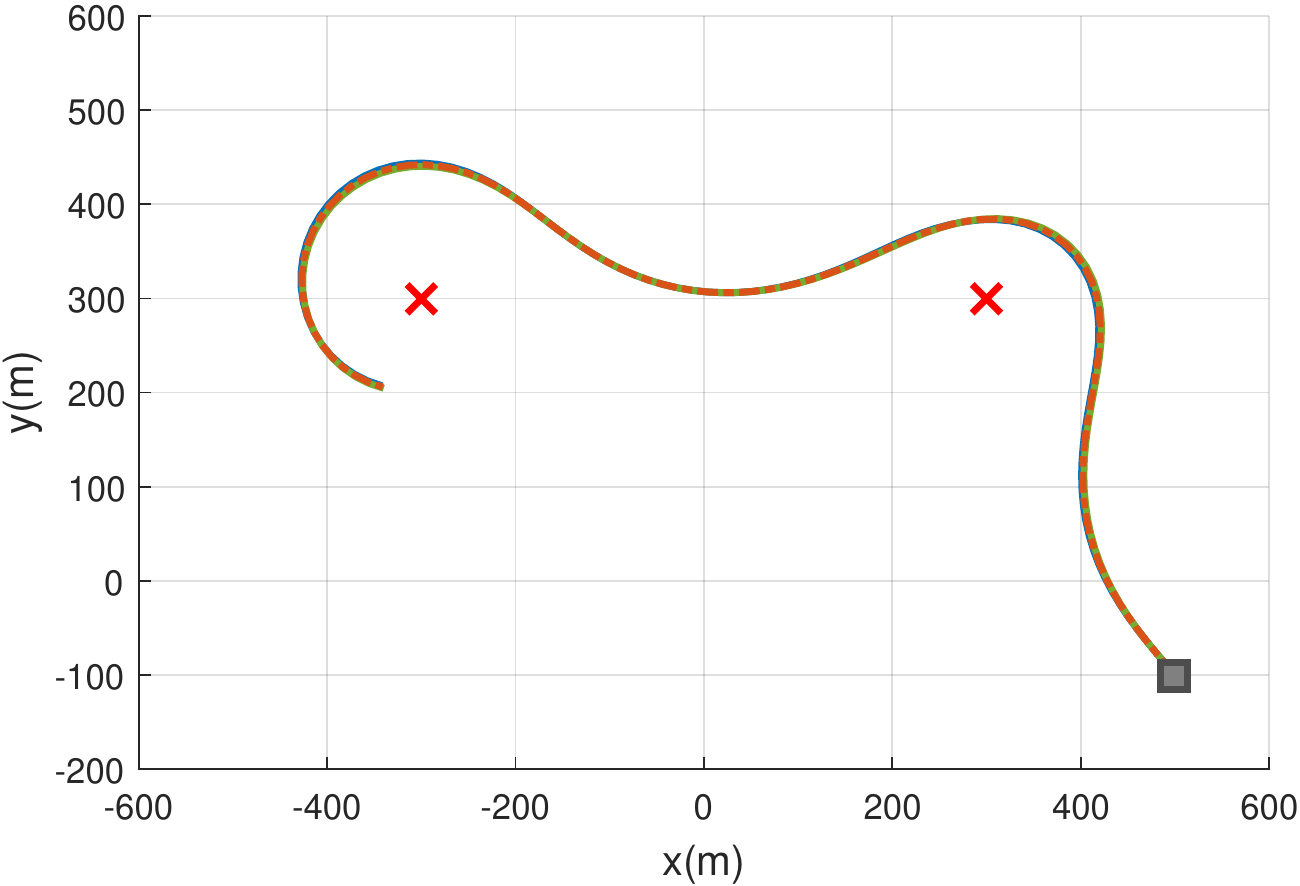}}\\
		%	\subfigure[Myopic trajectory planning]{
		%		\includegraphics[width=.65\columnwidth]{ex1_myopic}}
		\subfigure[Proposed algorithm]{
			\includegraphics[width=.65\columnwidth]{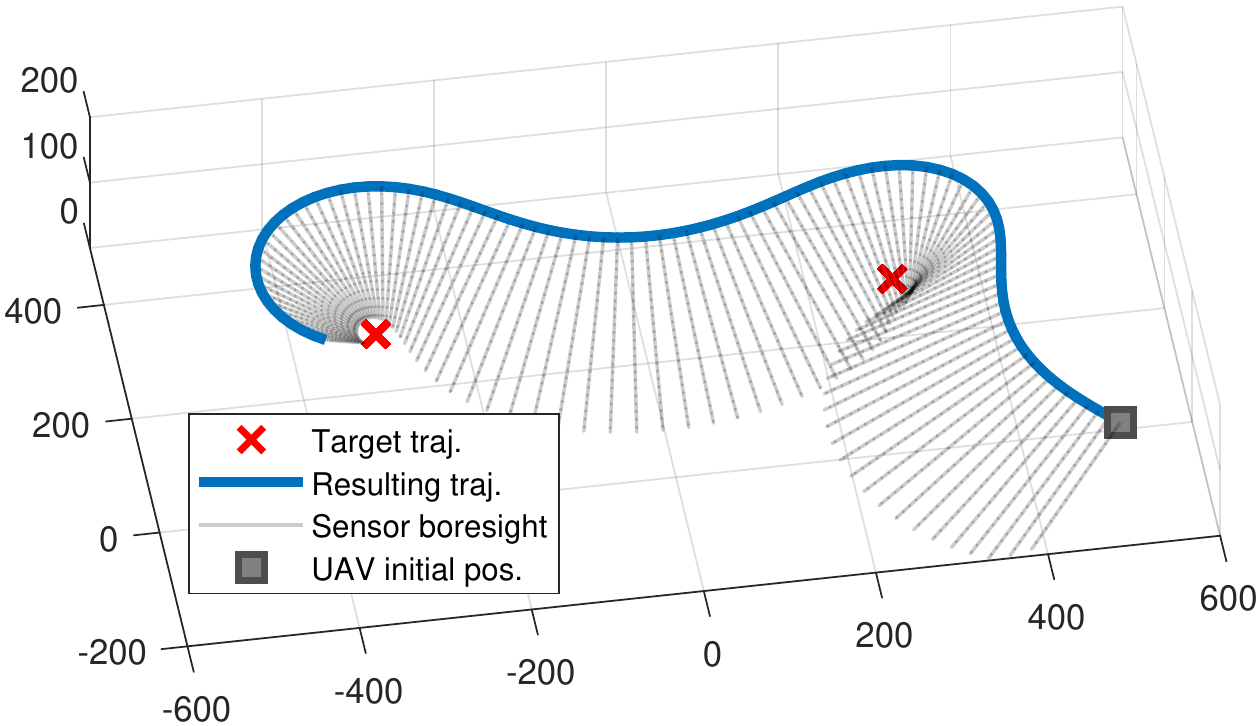}}
		\hspace*{0.5cm}
		\subfigure[Resulting variances of target estimations]{
			\includegraphics[width=.65\columnwidth]{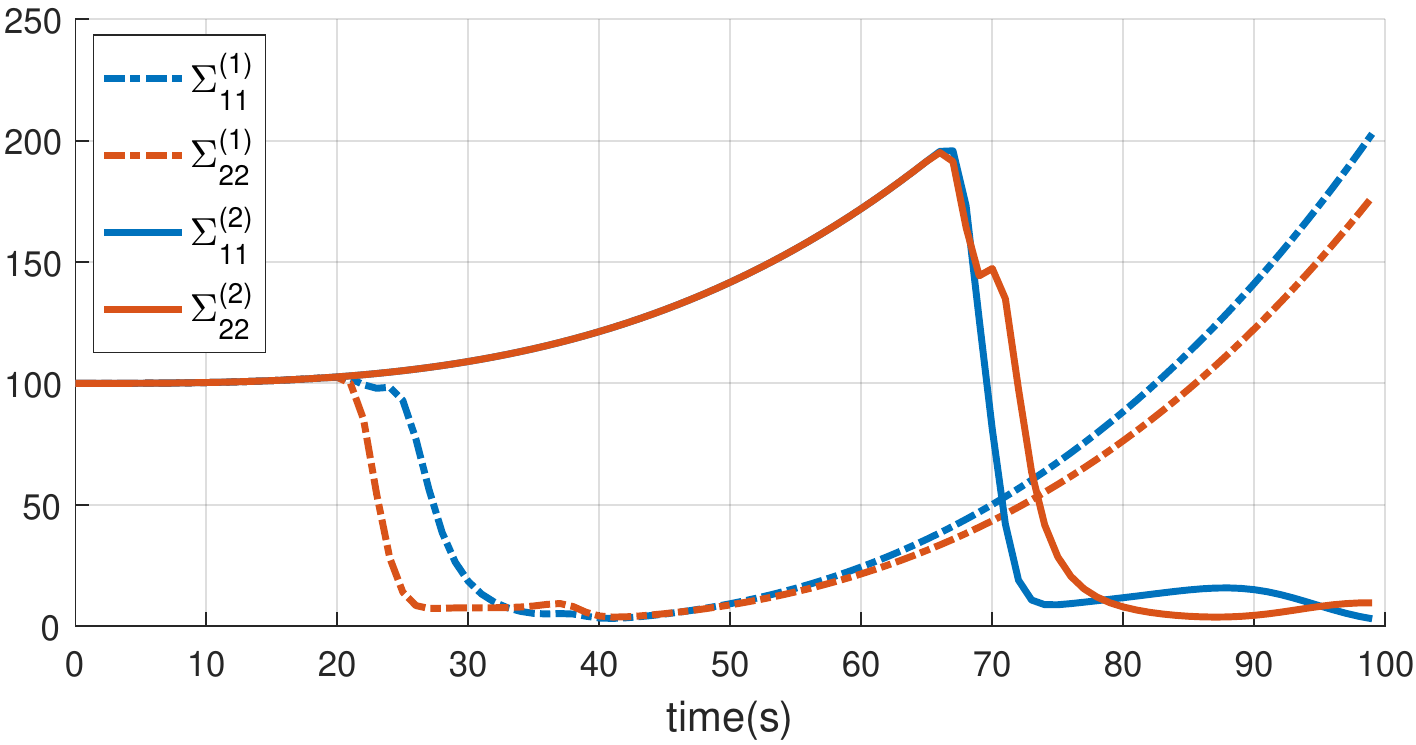}}
		%	\hspace*{0.5cm}	
		\caption{{(a)-(d) process of convergence and (e) final trajectory of the proposed non-myopic planning in a scenario with a single UAV and two stationary targets, and (f) the variances of the target estimations over the execution of the plan}}
		\label{fig: consensusf}
	}
\end{figure*} 

\begin{table*}[t]\centering{
		\caption{{Average uncertainty over 100 simulations for each algorithm in the single UAV scenarios.}}
		\begin{tabular}[t]{lccc} 
			\hline
			Number of targets & 2&3&4\\
			\hline
			Myopic trajectory planning \cite{oh2013coordinated}	 									&314.74&262.45&318.38\\	
			Non-myopic trajectory planning with initial guess $\textbf{u}_t = 0, \forall t$  &259.21&209.98&234.99\\
			Euclidean heuristic-based task-assignment \cite{luders2011information} + non-myopic trajectory planning 	&171.94&179.64&182.58\\
			Uncertainty-based task-assignment \cite{farmani2017scalable} + non-myopic trajectory planning				&167.63&172.67&173.88\\
			Proposed distributed trajectory optimization algorithm          				 							&124.52&150.76&149.11\\
			\hline
		\end{tabular} \label{tab:comp}
	}
\end{table*}%

We assume that the UAVs fly at different altitudes from each other to alleviate collision-avoidance constraints.
Throughout the entire simulations, the planning horizon, $T_p$, is set to 100 seconds, which is enough for the UAVs to cover all areas of interest with flying speed of $16m/s$.
Furthermore, we set $R = 100$ in \eqref{eq:main_traj_opt}, and  $\rho_{v} = \{0.02, 0.02, 500\times180/\pi, 500\times180/\pi\}$, $\rho_{u} = 500\times180/\pi$, $\alpha_{v}= 0.2\rho_{v}$ and $\alpha_{u}= 0.2\rho_{u}$ in \eqref{eq:lambda}. 
%Furthermore, we select $\rho_{v} = [2\text{E-2}, 2e-2, 2.86e4, 2.86e4]^{\top}$, $\rho_{u} = 2.86e4$, $\alpha_{v}= 0.2\rho_{v}$ and $\alpha_{u}= 0.2\rho_{u}$ in \eqref{eq:lambda}. 
It is important to set the penalty parameters, $\rho, \alpha$, properly for fast convergence of ADMM. 
%	Too large values of $\rho$ make the global common variable dominant in \eqref{eq:main_traj_opt}, and lead to less emphasis on reducing the uncertainty of the target. 
While they are chosen empirically in this work, many extensions of the distributed ADMM algorithm have been explored to adaptively select the penalty parameters. We would like to refer the readers to \cite{boyd2011distributed,song2016fast}.

For target motion model, we consider the motion on the XY-plane with a four-dimensional state vector of positions and velocities (in x and y axes) for a target. The matrix parameters of the linear stochastic dynamics in \eqref{eq:linear_stochastic} are set as:
%	A target motion is assumed by \eqref{eq:target_motion}, with the time interval between two successive measurements $d t = 0.1s$ and the constant diffusion strength $q=0.01$. 
%		Various models can serve as dynamics of the target. 
%	In this work, the motion of a target in a two-dimensional space is assumed to be modeled by the following matrices:
%\begin{equation}
%	\begin{split}
%		\textbf{x}_t^{(j)} = 
%		\begin{bmatrix}
%			x_t^{(j)} \\ y_t^{(j)} \\ v_{x,t}^{(j)} \\ v_{y,t}^{(j)}
%		\end{bmatrix},
%		A_t^{(j)} = 
%		\begin{bmatrix}
%			1 & 0 & d t & 0 \\ 0 & 1 & 0 & d t \\ 0 & 0 & 1 & 0 \\ 0 & 0 & 0 & 1
%		\end{bmatrix},\\
%		Q_t^{(j)} = q
%		\begin{bmatrix}
%			\frac{d t^{3}}{3} & 0 & \frac{d t^{2}}{2} & 0 \\ 0 & \frac{d t^{3}}{3} & 0 & \frac{d t^{2}}{2} \\ \frac{d t^{2}}{2} & 0 & d t & 0 \\ 0 & \frac{d t^{2}}{2} & 0 & d t
%		\end{bmatrix}, \label{eq:target_motion}
%	\end{split}
%\end{equation}
\begin{equation}
\begin{split}
A_t^{(j)} = 
\begin{bmatrix}
1 & 0 & d t & 0 \\ 0 & 1 & 0 & d t \\ 0 & 0 & 1 & 0 \\ 0 & 0 & 0 & 1
\end{bmatrix},
\Sigma_{w,t}^{(j)} = q
\begin{bmatrix}
\frac{d t^{3}}{3} & 0 & \frac{d t^{2}}{2} & 0 \\ 0 & \frac{d t^{3}}{3} & 0 & \frac{d t^{2}}{2} \\ \frac{d t^{2}}{2} & 0 & d t & 0 \\ 0 & \frac{d t^{2}}{2} & 0 & d t
\end{bmatrix}, \label{eq:target_motion}
\end{split}
\end{equation}
where $d t$ is the time interval between two successive measurements, and $q$ is the process noise intensity representing the strength of the deviations from the predicted motion by the dynamic model. When $q$ is small, this model represents a nearly constant velocity. For simulations, we set $d t = 0.1s$ and $q=0.01$. %\RE{\cite{charlish2011autonomous}}

For sensor platforms, we consider a set of fixed-wing  Unmanned Aerial Vehicles (UAVs). 
The dynamics of the UAVs is given by:	
\begin{equation}
\begin{split}
p_{x,t+1}^{(i)} &= p_{x,t}^{(i)} + V^{(i)} \, \cos \, \boldsymbol{\psi}_t^{(i)} d t,\\
p_{y,t+1}^{(i)} &= p_{y,t}^{(i)} + V^{(i)} \, \sin \, \boldsymbol{\psi}_t^{(i)} d t,\\
\boldsymbol{\psi}_{t+1}^{(i)} &= \boldsymbol{\psi}_{t}^{(i)} + g \, \tan \, \boldsymbol{\phi}_t^{(i)} d t\,/\,V^{(i)},\\
\boldsymbol{\phi}_{t+1}^{(i)} &= \boldsymbol{\phi}_{t}^{(i)} + u_t^{(i)} d t,
\end{split} \label{eq:uav_dyn}
\end{equation}		
where $[p_{x,t}^{(i)}, p_{y,t}^{(i)}]$, $V^{(i)}$, $\psi_t^{(i)}$ and $\phi_t^{(i)}$ are the position, speed, heading angle and bank angle of 
$i$-th UAV at time $t$, respectively. Gravity acceleration is denoted as $g$.
%	Sensors are assumed to  measure the kinematic information about the targets relative to the sensor, itself. 

The sensors are mounted on the left or right sides of the UAVs and are aimed in a $30$-degree downward direction as shown in Fig.~\ref{fig: sensor_geo}.
%The sensors are supposed to be mounted at left or right side of the UAVs and aim down
%The sensor are mounted on the left or right side of the UAVs and is aimed at a $30$-degree downward direction, with
%The parameters of the measurement model in \eqref{eq:sensor_model} are $\alpha = 10$, $\gamma = 8ln2$ and $\beta=0$.
Each sensor measures the signal-to-noise ratio (SNR) \cite{skolnik1970radar}:	
\begin{equation}
h^{(i)}(\textbf{x}_t^{(j)},p_t^{(i)}) = \frac {\alpha \, e^{ -\gamma \,\big(\theta_t^{(i,j)}\big)^2}}{\big(r_t^{(i,j)}\big)+\beta}, 
\label{eq:sensor_model}
\end{equation}			
where $\alpha$, $\beta$, and $\gamma$ are selected to model the SNR of the sensor.
$r_t^{(i,j)}$ and $\theta_t^{(i,j)}$ represent the distance between the sensor and the target as well as the angle between the sensor's boresight and the direction from the sensor to the target, respectively.
When $\gamma$ is set to zero, the sensor measures the quasi-distance, which is a commonly used model \cite{choi2009roles,williams2007approximate,hoffmann2010mobile}.
We set $\alpha = 10$, $\gamma = 8ln2$ and $\beta=0$.
%	The sensors are assumed to be mounted at the left or the right side of the UAVs and aim down (Fig. \ref{fig: sensor_geo}).

%\subsection{Case 1: Single UAV}

\subsection{Process of Convergence}

First, we consider a toy example in which a single UAV tracks two targets with zero initial velocities to explore the convergence process of the proposed distributed trajectory optimization algorithm.
%	Fig. \ref{fig: consensusf}(a) shows the resulting trajectory when one-step myopic trajectory planning is repeated.
Figs. \ref{fig: consensusf} (a-d) present the snapshots of the optimization process.
In the beginning, the algorithm generates an initial guess (trajectory) of the UAV for each target independently, without any high-level decisions to consider both targets simultaneously. In the early phase, note that the consensus trajectory largely deviates from the distributively optimized one for each target. 
As the number of iterations increases, the deviation gets smaller and eventually vanishes.
This convergence is enabled as each subproblem reflects the result of the other subproblem from the previous iteration through the Lagrange multipliers in \eqref{eq:main_traj_opt}. By repeating this interaction, the algorithm provides a non-myopic trajectory planning result that sequentially tracks two targets, as shown in Fig. \ref{fig: consensusf}(e).

In addition to observing the convergence, we evaluate the performance of the converged trajectory in reducing the uncertainty of each target. As the components of the posterior uncertainties after executing the plan, the variances of the target state (position) estimations are logged along the time as shown in \ref{fig: consensusf}(f). The variances of each target are effectively reduced when the target is in the sensing range, while they increase gradually due to the stochastic motion of the target when it is in sensing holes.
%	The consensus trajectory is obtained through the optimization results of each subproblem, which affects the optimization in the next iteration through the Lagrange multipliers;
%	that is, the optimization results of each subproblem reflect the optimization process of the other subproblems of the previous iteration.
%	This procedure is repeated, and finally we obtain the converged solution.
%that tracks two targets sequentially

\begin{figure*}[t]	
	{
		\begin{minipage}{0.33\textwidth}	
			\centering
			\hspace*{0cm}%
			\subfigure[Myopic trajectory planning]{
				\includegraphics[width=1\columnwidth]{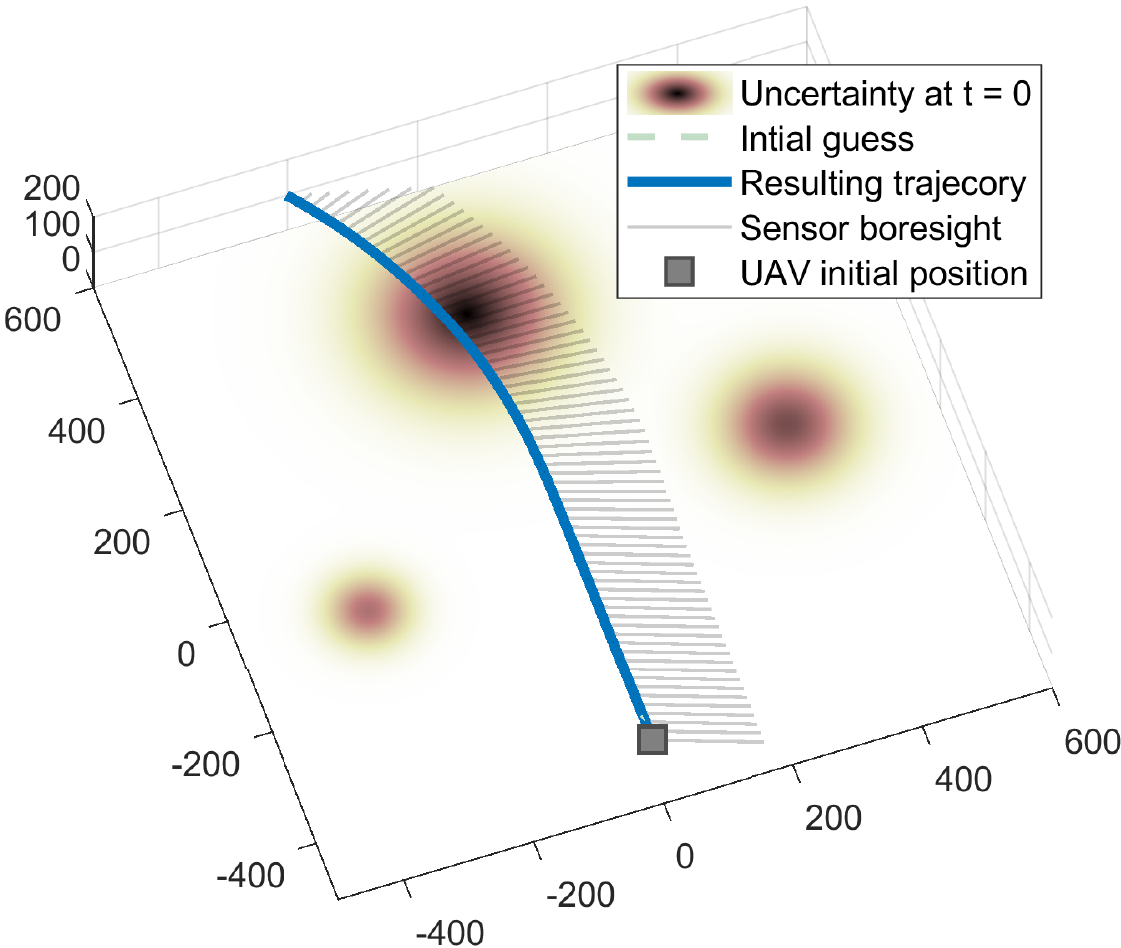}}
			%	\hspace*{1cm}%
			\subfigure[Non-myopic trajectory planning with initial guess $\textbf{u}_t = 0, \forall t$]{
				\includegraphics[width=1\columnwidth]{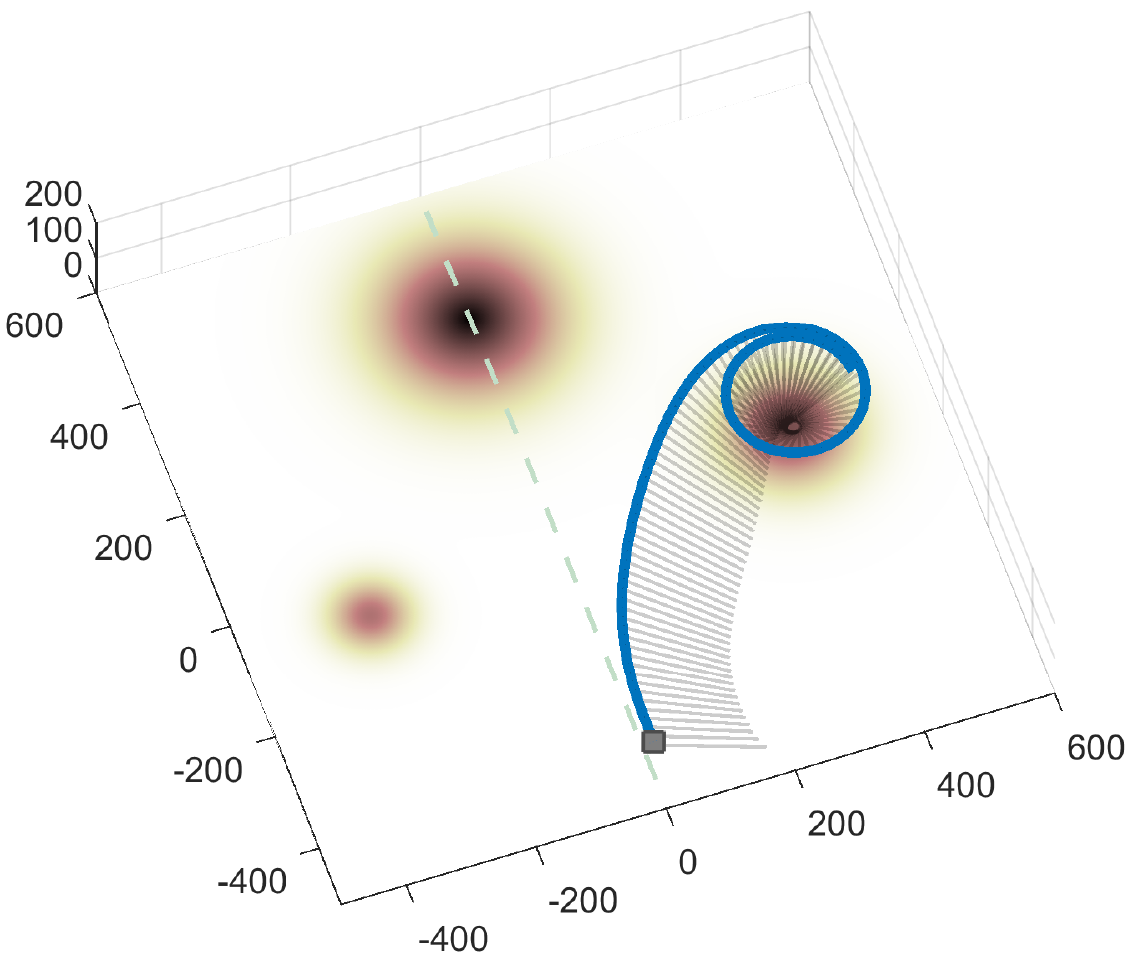}}
		\end{minipage}
		\hspace*{0.3cm}%
		\begin{minipage}{0.33\textwidth}	
			\centering
			\subfigure[Euclidean heuristic-based task-assignment + non-myopic trajectory planning]{
				\includegraphics[width=1\columnwidth]{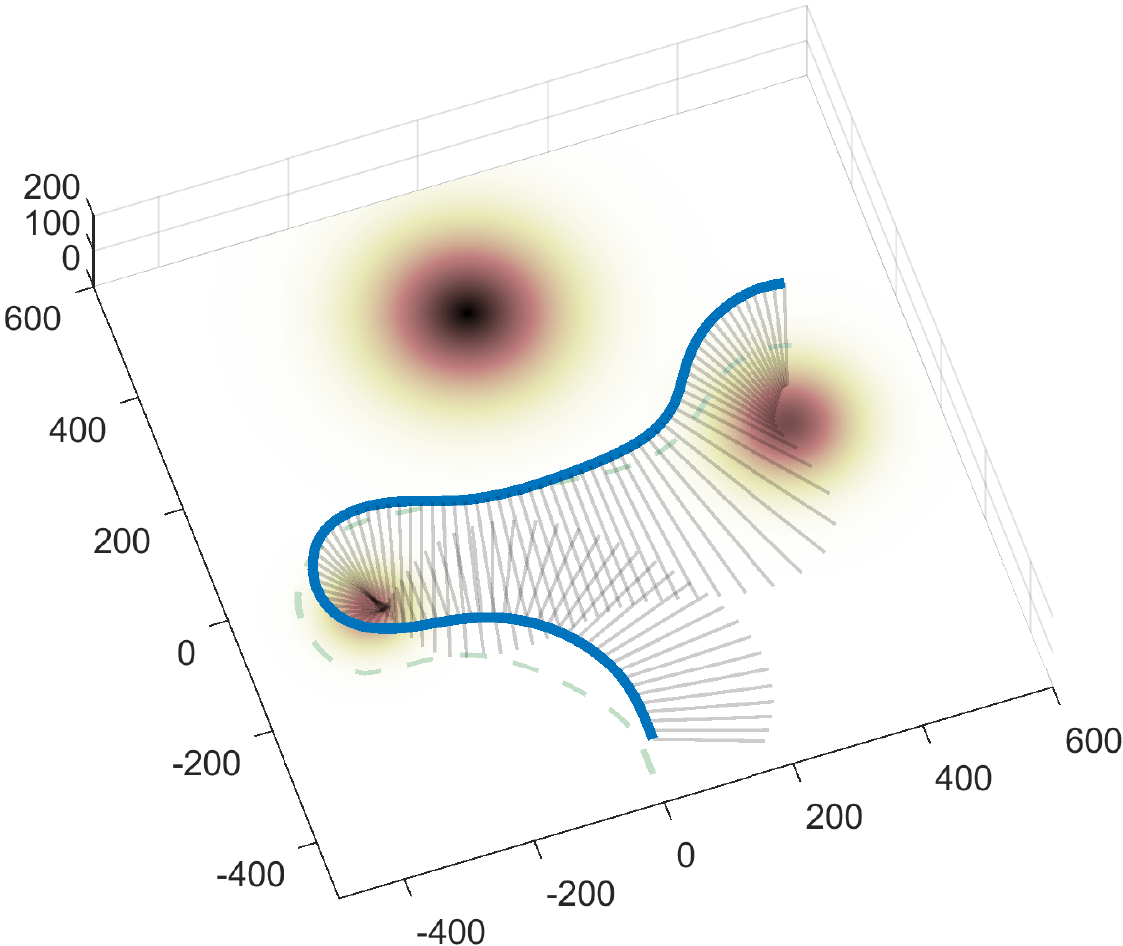}}	
			%	\hspace*{1cm}%
			\subfigure[Uncertainty-based task-assignment + non-myopic planning]{
				\includegraphics[width=1\columnwidth]{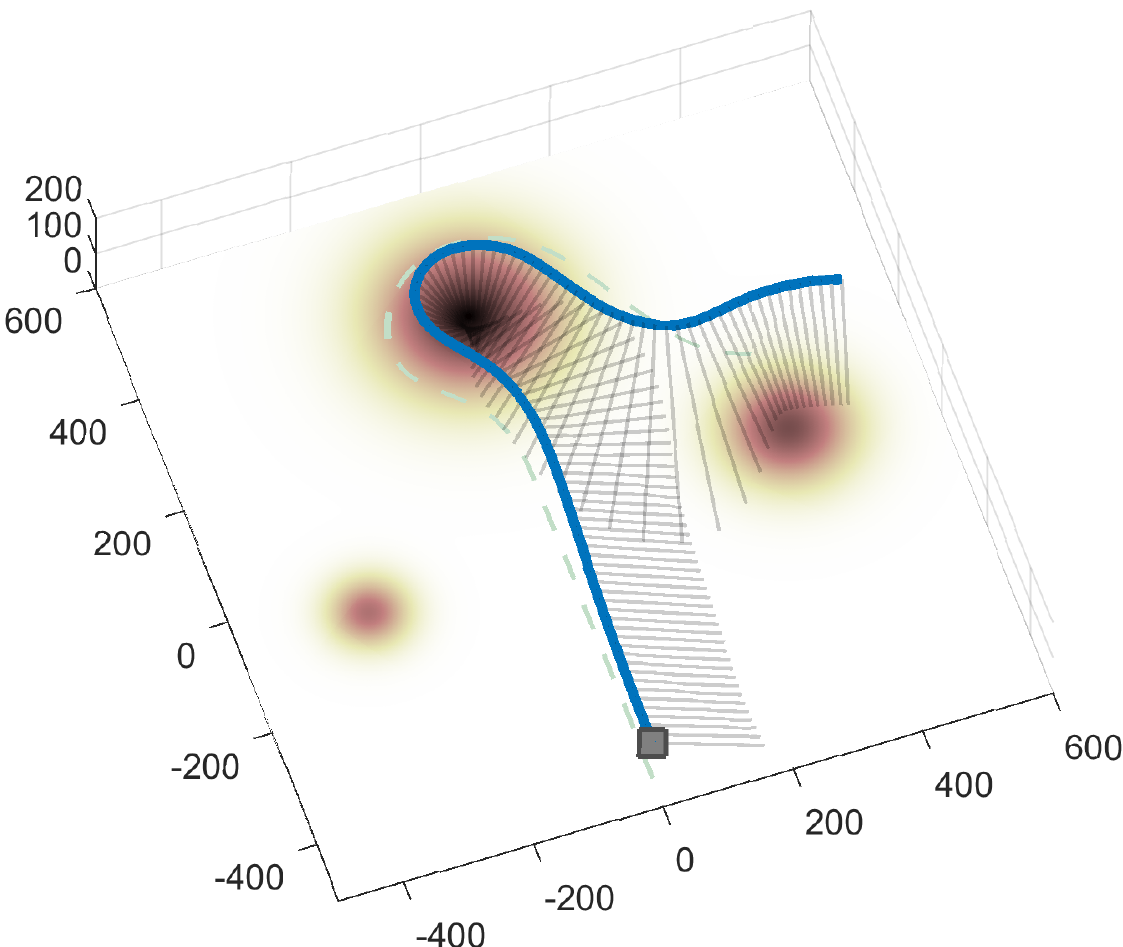}}
		\end{minipage}%
		\hspace*{0.3cm}%
		\begin{minipage}{0.33\textwidth}	
			\centering
			\subfigure[Proposed algorithm]{
				\includegraphics[width=1\columnwidth]{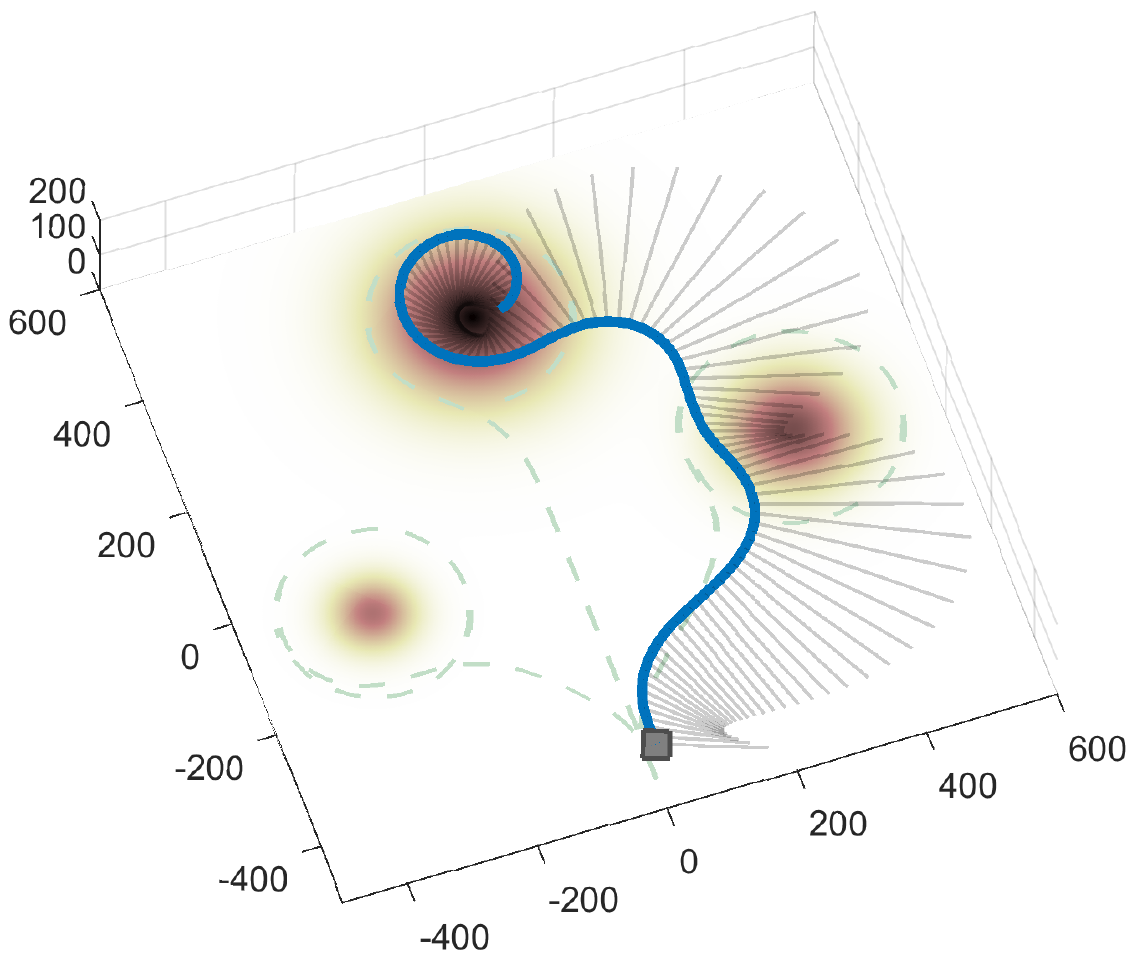}}
		\end{minipage}%
	}
	\caption{{Comparison of existing approaches and the proposed algorithm in a scenario with a single UAV and three targets}}
	\label{fig:comp_naive}
\end{figure*}
\begin{table*}[t]
	\centering{
		\caption{{Average result over 1000 simulations for each algorithm in the single UAV scenario.}}
		\begin{tabular}[t]{lccc} 
			\hline
			&Uncertainty&Posterior Uncertainty& Localization RMSE [m]\\
			\hline
			Myopic trajectory planning \cite{oh2013coordinated}	 									&252.15&252.47&8.4400\\	
			Non-myopic trajectory planning with initial guess $\textbf{u}_t = 0, \forall t$  &204.01&204.23&7.1812\\
			Euclidean heuristic-based task-assignment \cite{luders2011information} + non-myopic trajectory planning 	&194.56&194.32&6.8059\\
			Uncertainty-based task-assignment \cite{farmani2017scalable} + non-myopic trajectory planning				&169.23&169.50&6.7532\\
			Proposed distributed trajectory optimization algorithm          				 							&147.39&146.95&6.5595\\
			\hline
		\end{tabular} \label{tab:comp_rmse}
	}
\end{table*}%
\begin{table*}[h] 
	\centering{
		\caption{{Average uncertainty over 100 simulations for each algorithm in two-UAV scenarios.}}
		\begin{tabular}[t]{lccc} 
			\hline
			Number of targets & 10&15\\
			\hline
			Target clustering \& Euclidean heuristic-based task-assignment \cite{luders2011information} + non-myopic trajectory planning &557.01 &498.53\\
			Target clustering \& uncertainty-based task-assignment \cite{farmani2017scalable} + non-myopic trajectory planning			 &566.55 &501.75\\
			Proposed distributed trajectory optimization algorithm          				 											 &389.59 &377.06\\
			\hline
		\end{tabular} \label{tab:comp2}}
\end{table*}%

%\subsection{Comparison to Existing Works} % with Heuristic Task-assignment
\subsection{Single UAV tracking a Few Targets}

In the second example, we compare the proposed algorithm with existing approaches under scenarios where one UAV tracks a few targets with zero initial velocities. 
%The comparison is performed for three cases where the number of targets to be tracked is different, with one UAV.
%	The targets are assumed to be stationary, 
The initial positions of the UAV and the targets on the XY-plane are randomly (uniformly) generated within a $1200m \times 1200m$ area.
The targets are on the ground level and the UAV is flying at a fixed altitude of $100m$.
The initial covariance matrices of the target estimations are also randomly generated as diagonal matrices of which diagonals are picked uniformly between 10 and 100.

As the existing works \cite{farmani2017scalable,luders2011information} on multi-target tracking tasks are myopic approaches, we extend them to non-myopic ones. We adopt their task-assignment algorithms to produce initial guesses on the original undistributed problem \eqref{eq:original_cost} and solve it through belief space iLQG as in our algorithm.
The algorithms assign clustered targets to the closest mobile sensor (w.r.t. the center of the cluster) and determine the priorities of the clusters when they assigned to the same sensor.
In \cite{farmani2017scalable}, the priority is determined as the sum of the uncertainties (at $t=0$) of the targets in each cluster, while \cite{luders2011information} uses the Euclidean distance of each cluster from its assigned mobile sensor and the number of targets in the cluster. 
Initial guesses are generated in a similar way as described in Section \ref{subsection:initial_guess}, as shown as the dotted lines in Figs. \ref{fig:comp_naive}(c-d), so that the UAV can visit the clusters (single target for each cluster in this case) sequentially based on the task-assignment results.
Additionally, we tested two more undistributed algorithms. First, we apply a myopic algorithm \cite{oh2013coordinated} to \eqref{eq:original_cost} repeatedly to cover the planning horizon with single-step plans. The other is solving \eqref{eq:original_cost} with a naive initial guess, $\textbf{u}_t = 0 \;\forall t$.
\begin{figure*}[t] 
	\centering
	{
		\hspace*{0.cm}%
		\subfigure[Target clustering \& Euclidean heuristic/uncertainty-based task-assignment + non-myopic trajectory planning]{
			\includegraphics[width=.855\columnwidth]{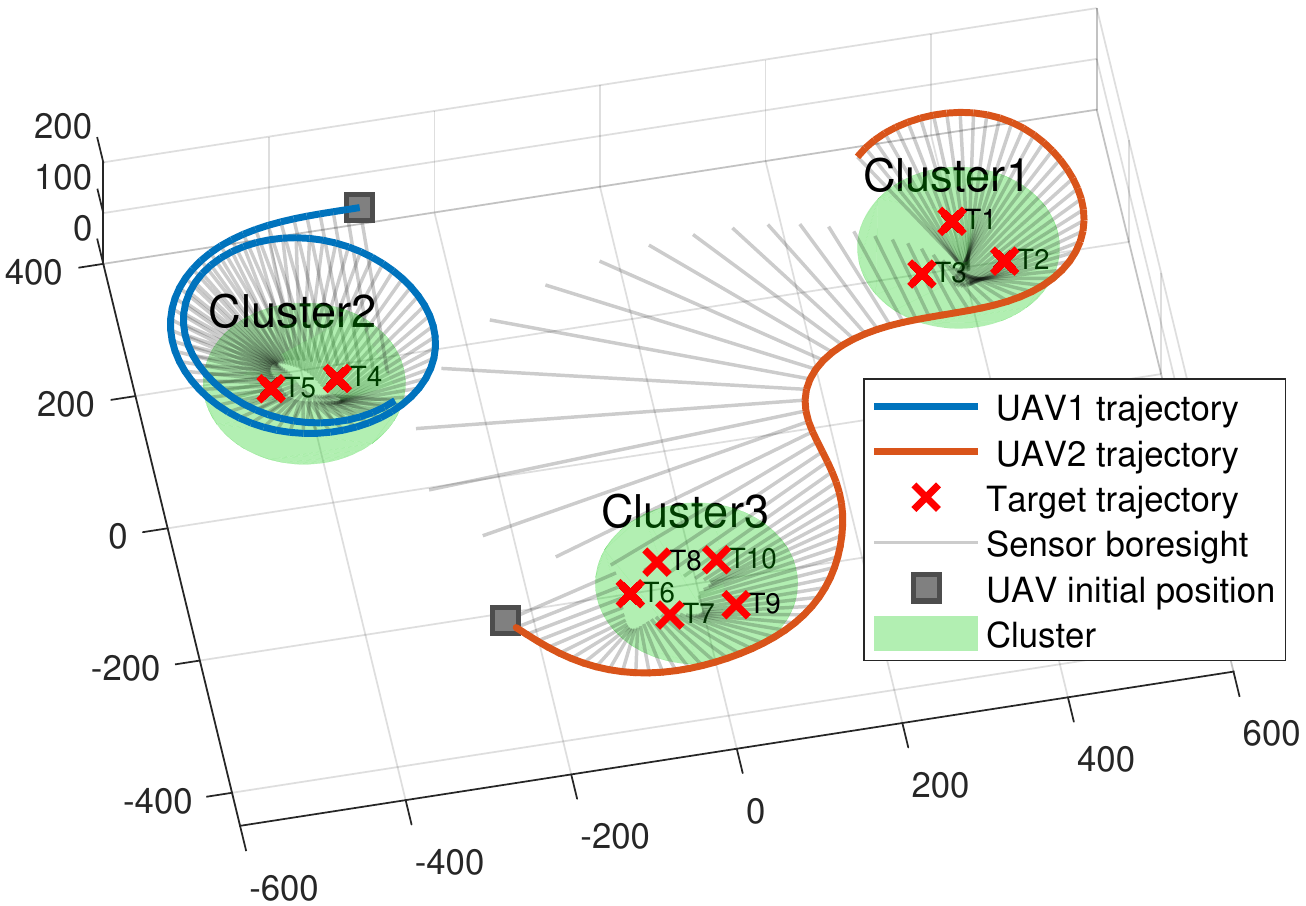}}	
		\hspace*{1.cm}%
		\subfigure[Proposed algorithm]{
			\includegraphics[width=.855\columnwidth]{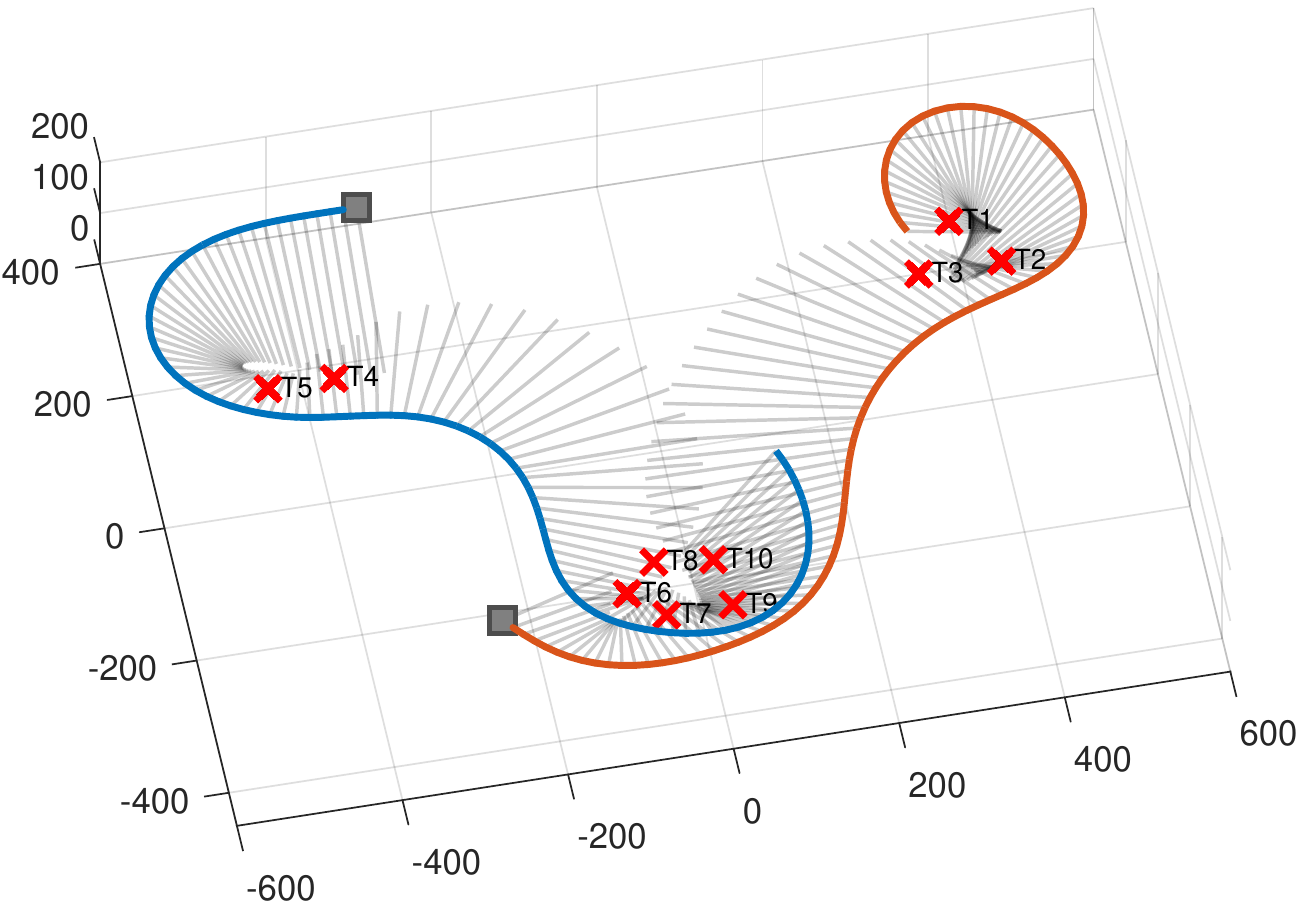}}
		\hspace*{0.cm}%
	}
	\caption{Comparison of two-phase approaches and the proposed algorithm for a scenario with two UAVs and 10 targets}
	\label{fig:multitarget}
\end{figure*}

\begin{table*}[h] 
	\centering{
		\caption{{Average result over 1000 simulations for each algorithm in the two-UAV scenario.}}
		\begin{tabular}[t]{lccc} 
			\hline
			&Uncertainty&Posterior Uncertainty& Localization RMSE [m]\\
			\hline
			Clustering based heuristic task-assignment \cite{luders2011information,farmani2017scalable} + non-myopic trajectory planning &584.72 &576.08 &15.8290\\
			Proposed distributed trajectory optimization algorithm          				 											 &349.22 &357.95 &13.3126\\
			\hline
		\end{tabular} \label{tab:comp2_rmse}}
\end{table*}%

Table \ref{tab:comp} summarizes the average result over the 100 randomly generated topologies for each number of targets. To evaluate the planned trajectory, we compare the uncertainty of target estimation, which is the primary objective to minimize in our problem, averaged over targets, timesteps, and topologies. 
Since the uncertainty has a meaning of the average variance of target estimation, we can compare the values approximately in a quadratic scale. Our algorithm shows the best performance among the tested algorithms. To analyze the details, we investigate a single instance of the randomly generated topologies in depth.

Fig. \ref{fig:comp_naive} shows the result of each algorithm in a scenario of single UAV tracking three targets.
As shown in the Fig. \ref{fig:comp_naive}(a), the myopic algorithm just plans a trajectory that minimizes only the control efforts until any target gets in the sensing range of the UAV. This is because the UAV cannot reduce the uncertainty without observing the targets.
%	as the current control effort is not propagated to future time steps. That is, current control efforts do not help to improve future estimation performance outcomes.
%	The sensor has a limited field of view, and all targets are located out of view at the planning time and one-time step later.\cite{liu2003multi,lee2018potentialaccess} Thus, the current control effort is not propagated in the future time step, and an optimal solution is obtained that minimizes the control effort only.
In Fig. \ref{fig:comp_naive}(b), the resulting trajectory tracks only a single target. This shows that it could be stuck in a bad local optimum to perform non-myopic trajectory optimization without proper decision making. 
The two-phase approaches in Figs. \ref{fig:comp_naive}(c-d) show better trajectories so that they sense more targets. However, the trajectories include more portion of idle time without observing a target than the proposed algorithm in Fig. \ref{fig:comp_naive}(e).

To compare the results numerically, we examine their average performance over 1000 randomly generated instances on the topology of Fig. \ref{fig:comp_naive}. As shown in Table \ref{tab:comp_rmse}, we evaluate posterior uncertainty and root mean square error (RMSE) of target localization in addition to uncertainty. 
Even though the topology was simple and easy to have intuitive decision making, the proposed algorithm still outperforms the others. 
While the two-phase approaches define new approximate/heuristic cost functions for the task-assignment and that decision making considers only the current status of the sensor networks, the proposed algorithm incorporates the decision making into the optimization process while considering the mobility of the sensor platform and the target state through the distributed formulation.

\subsection{Multiple UAVs tracking Numerous Targets} \label{sec:sim_edge}

The third example considers the situation in which two UAVs track numerous targets with zero initial velocities.
To compare the proposed algorithm to the two-phase approaches \cite{farmani2017scalable,luders2011information} involving target clustering, the topologies of the targets and UAVs are randomly generated based on three-cluster structure. (Choose centers of the clusters and the number and positions of the targets for each cluster randomly.)
%	For the execution of our algorithm, the initial guess for each pair of targets and UAVs is generated in the same manner as in the previous examples.
%	The algorithm is compared with the two-phase approaches \cite{farmani2017scalable,luders2011information} involving target clustering.
For the clustering algorithm, density-based spatial clustering of applications with noise (DBSCAN) \cite{ester1996density} is used. As parameters of DBSCAN, we set the minimum number of targets for a single cluster and the maximum radius of a cluster to 2 and 150$m$, respectively.
Table \ref{tab:comp2} summarizes the results. While the two-phase approaches show similar average uncertainty values, the proposed algorithm outperforms them.

We further investigate an instance of the randomly generated topologies to analyze the performance gap in detail.
Fig. \ref{fig:multitarget} shows the resulting trajectories of the two-phase approaches and the proposed algorithm in a scenario with 10 targets.
The Euclidean heuristic and uncertainty-based task-assignments provide the same initial guesses, where each cluster is assigned to a single UAV.
However, the proposed algorithm automatically generates the trajectories for the two UAVs to visit the same cluster. Table \ref{tab:comp2_rmse} shows that the trajectories outperform those from the two-phase approaches. 
Even in this well-clustered topology, the proposed algorithm plans trajectories more optimized for the mobile sensors to visit the target clusters without the aid of clustering and task-assignment algorithms. 
%Allowing two UAVs to visit the same cluster may be better in terms of estimation performance, but the two-phase approaches do not allow it.

%	Even in well clustered topology, it is hard to make a decision whether a cluster should be assigned dominantly or not.

\begin{figure*}[h] 
	\centering{
		\hspace*{0.cm}%
		\subfigure[$\epsilon$ = 0 (not applied)]{
			\includegraphics[width=.6\columnwidth]{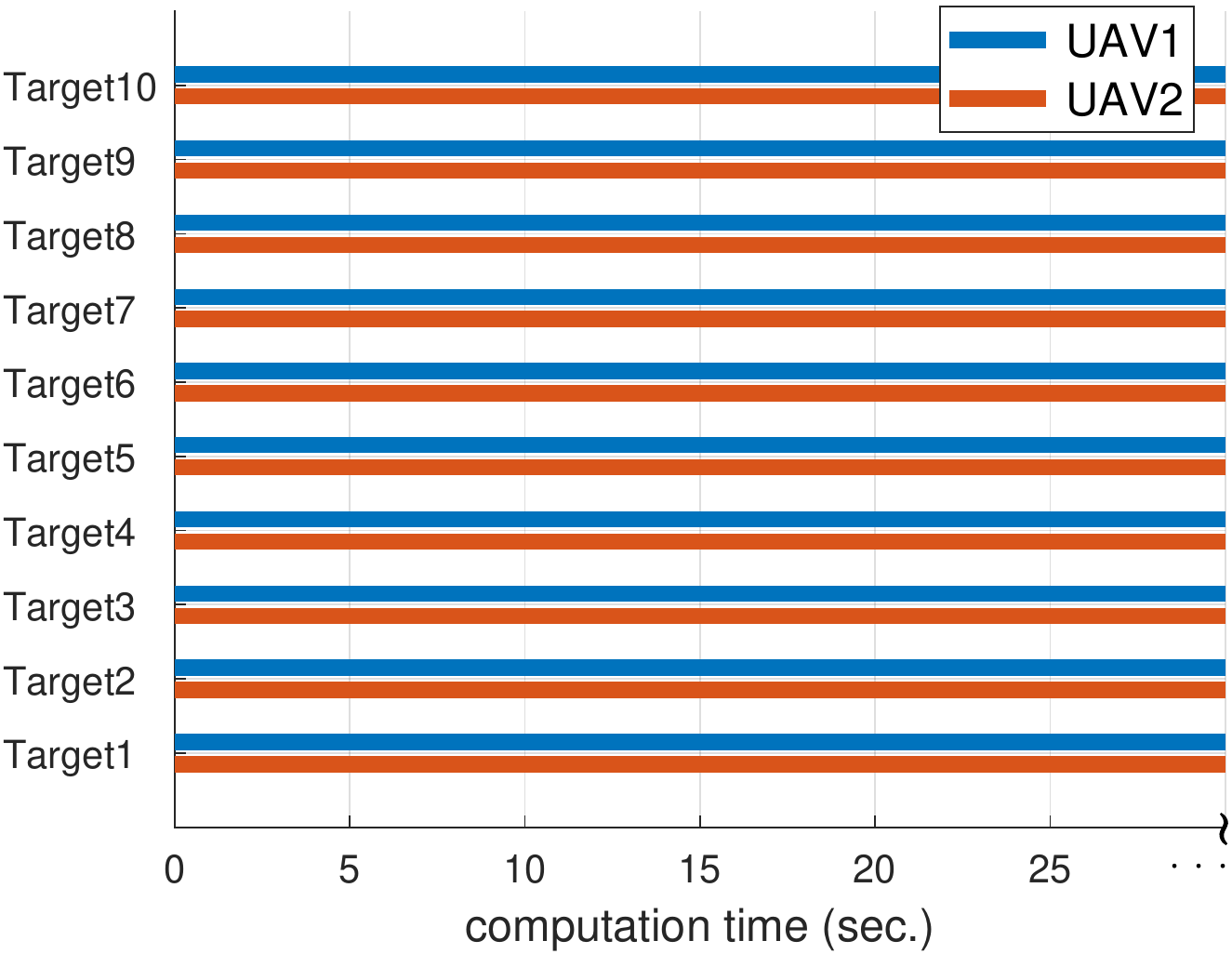}}	
		\subfigure[$\epsilon$ = 2000]{
			\includegraphics[width=.6\columnwidth]{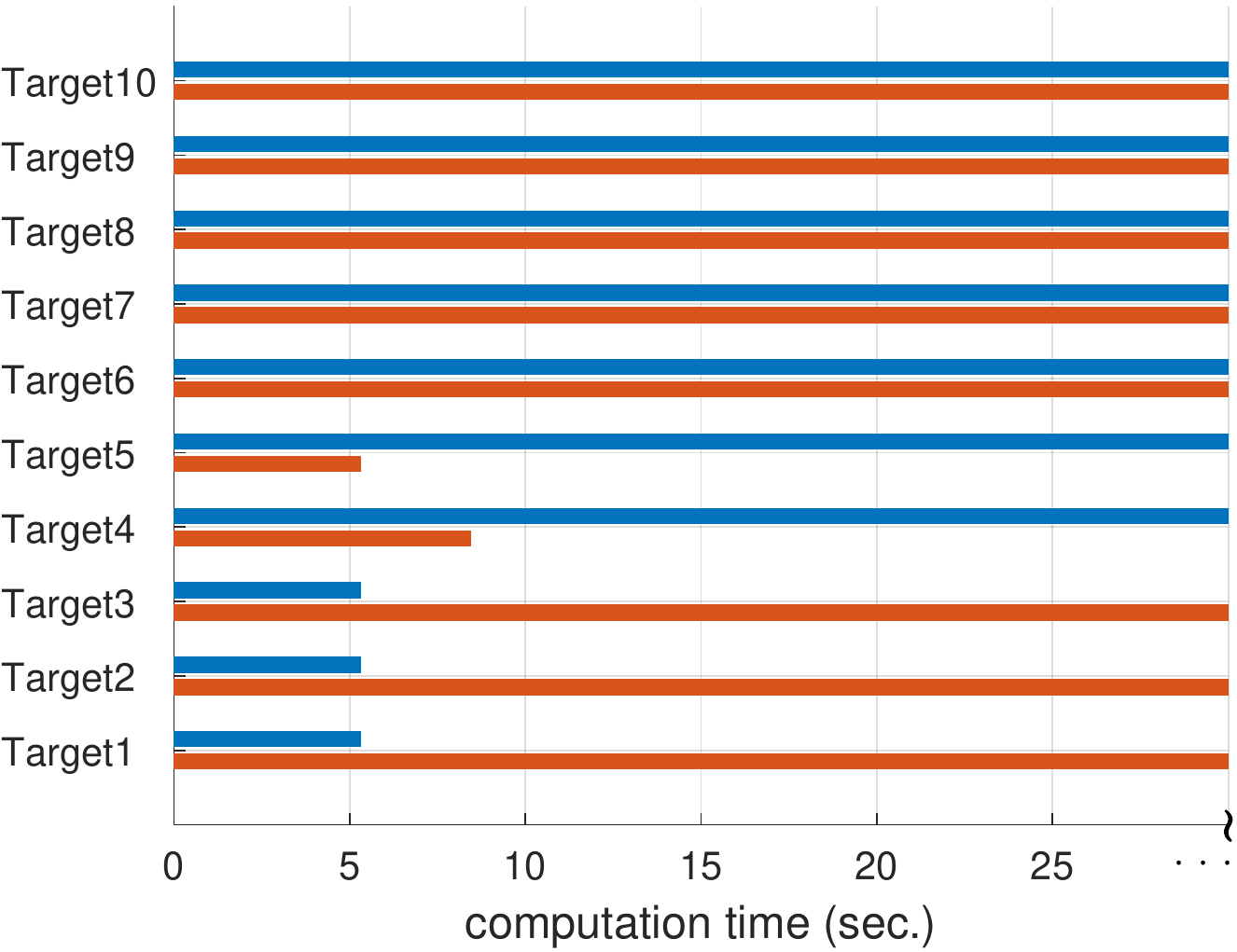}}
		\hspace*{0.cm}%
		\subfigure[$\epsilon$ = 6000]{
			\includegraphics[width=.6\columnwidth]{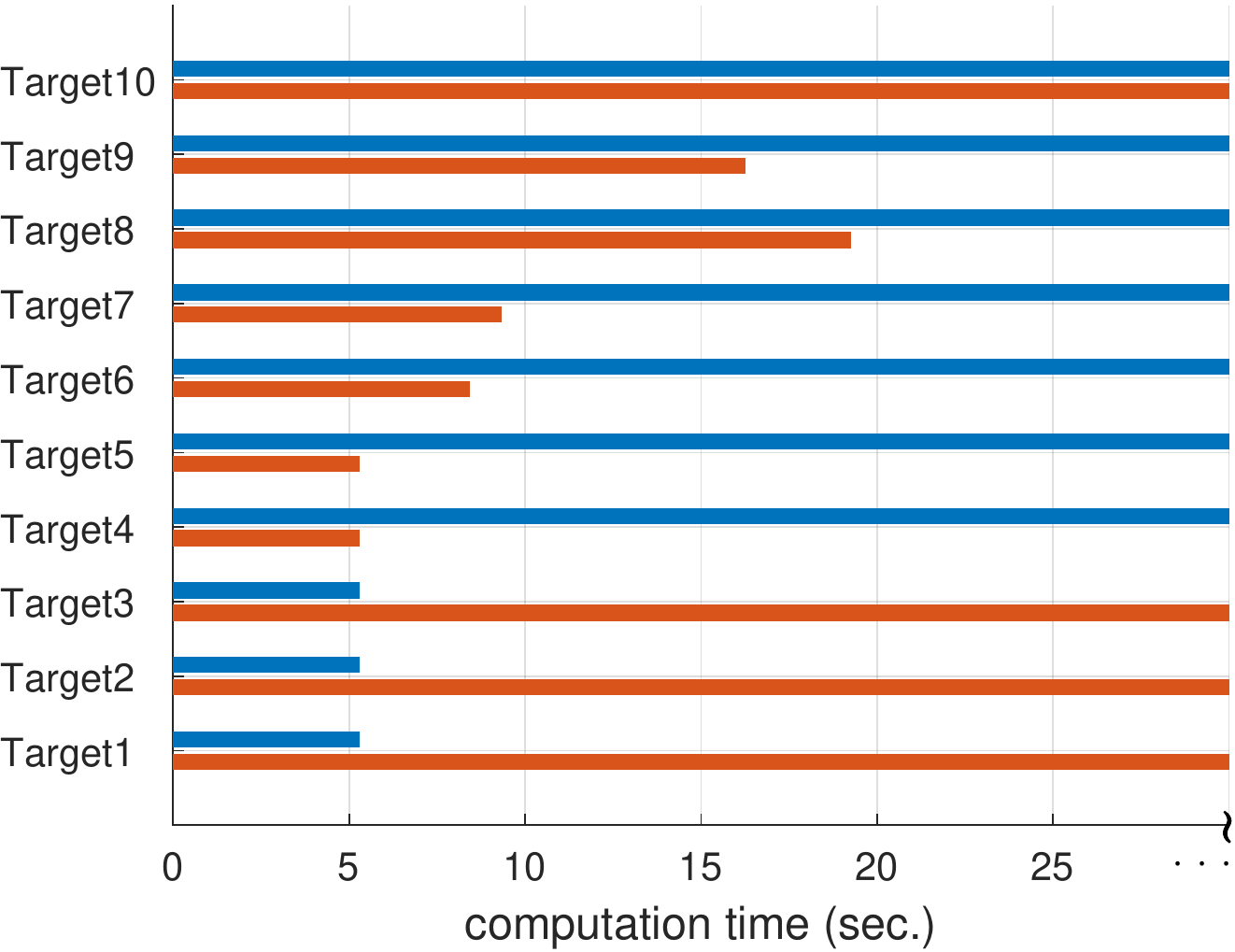}}\\
		\subfigure[Cost comparison w.r.t. $\epsilon$ value]{
			\includegraphics[width=.6\columnwidth]{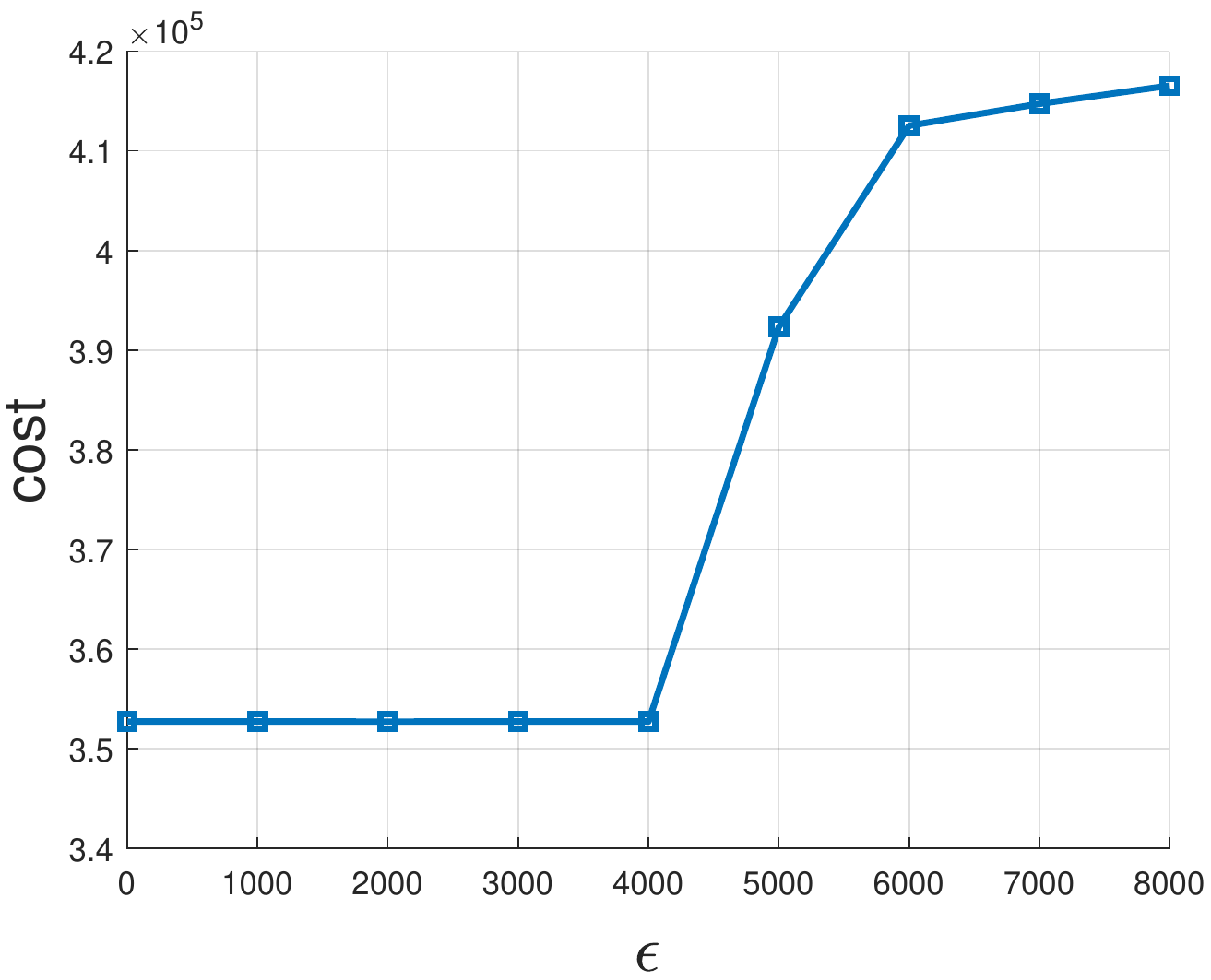}}		
		%	\hspace*{1.cm}%
		\subfigure[Results without parallel computing]{
			\includegraphics[width=.58\columnwidth]{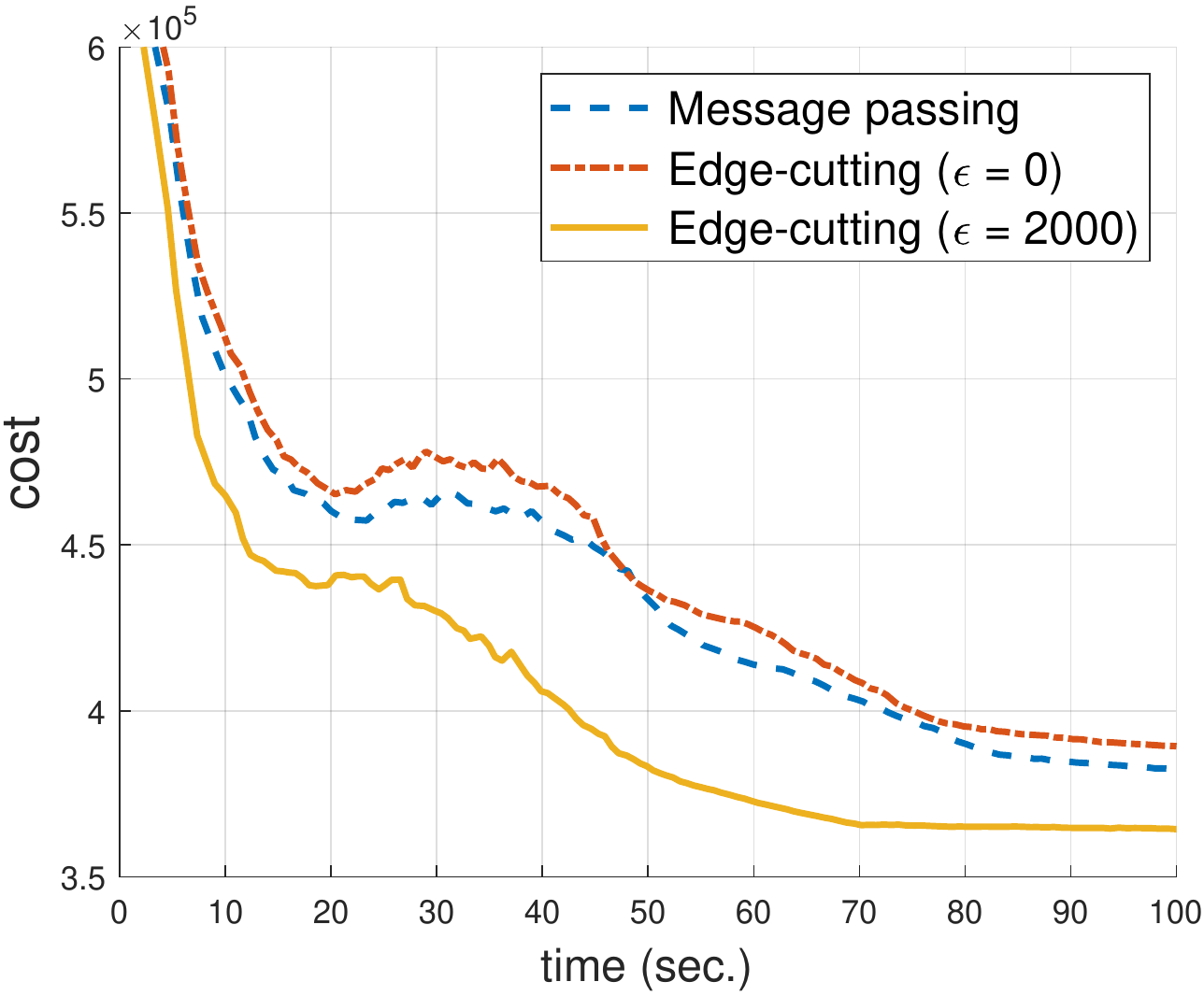}}		
		\subfigure[Results with parallel computing]{
			\includegraphics[width=.58\columnwidth]{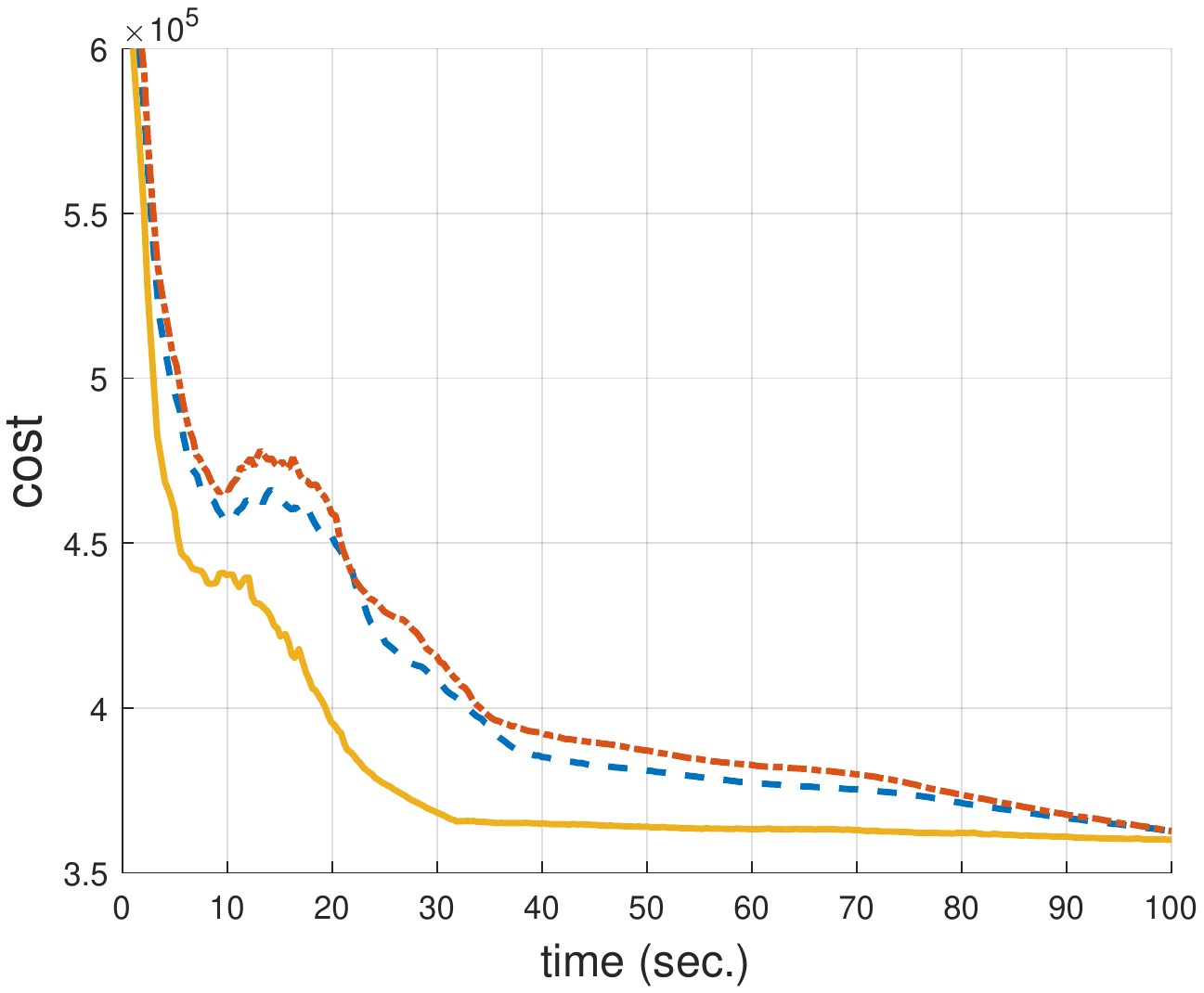}}		
		\caption{{(a)-(c) Cutting time analysis for the edge-cutting method with three different $\epsilon$ values (Each solid line corresponds to an edge and indicates that the UAV is considering the  target in the trajectory optimization process), (d) cost comparison for different $\epsilon$ values and (e)-(f) comparison of the computation times}}
		\label{fig:cutting_edge}}
	
	%		\centering
	%		\subfigure[time = 32 sec.]{
	%			\includegraphics[width=.65\columnwidth]{ex4_replan32_n.pdf}}
	%		\subfigure[time = 96 sec.]{
	%			\includegraphics[width=.65\columnwidth]{ex4_replan96_n.pdf}}
	%		\subfigure[time = 160 sec.]{
	%			\includegraphics[width=.65\columnwidth]{ex4_replan160_n.pdf}}
	%		\subfigure[time = 256 sec.]{
	%			\includegraphics[width=.65\columnwidth]{ex4_replan256_n.pdf}}
	%		\subfigure[Position error]{
	%			\includegraphics[width=.65\columnwidth]{ex4_error_position}}
	%		\subfigure[Velocity error]{
	%			\includegraphics[width=.65\columnwidth]{ex4_error_velocity}}
	%		\caption{(a)-(d) Snapshots of the current plan (solid line) and future plan (dashed line) for each mobile sensor in the replanning process scenario with two UAVs, one moving target and two stationary targets, and (e)-(f) particle filter estimation results}
	%		\label{fig: replanf}
\end{figure*}

\subsection{Evaluation of Edge-Cutting Method}

We evaluate the edge-cutting method for the scenario in Fig. \ref{fig:multitarget}(b). Varying the $\epsilon$ values, the time when each edge is cut is analyzed in Fig. \ref{fig:cutting_edge}(a-c). As expected, larger $\epsilon$ implies faster cutting time since it plays a role of a weaker threshold. However, as shown in Fig. \ref{fig:cutting_edge}(d), too large $\epsilon$ values make the solution of the trajectory optimization converge to a bad local optimum with a larger cost. This is because too easily cut edges change the original problem significantly beyond the degree of proper approximation.  
%Fig. \ref{fig:cutting_edge} shows the effectiveness of the proposed edge-cutting method.
When the epsilon value is set appropriately, not too large, the edge-cutting method produces the same result as it is not applied ($\epsilon = 0$) with effectively reduced dimensions of the optimization.

%	Figs. \ref{fig:cutting_edge}(a-c) show that the edges are maintained and cut for different $\epsilon$ values.
%	If the set $\epsilon$ value is not large, the same result can be obtained as in cases where the edge-cutting method is not applied ($\epsilon = 0$), as shown in Fig. \ref{fig:cutting_edge}(d) and Fig. \ref{fig:multitarget}(b). 
%	Specifically, when the epsilon value is set appropriately, trajectory optimization associated with the cut edges is excluded, reducing the dimensions of the subproblems.
%Thus, the edge-cutting method reduces the computation time of the proposed trajectory optimization algorithm.

In Figs. \ref{fig:cutting_edge}(e-f), the actual computation time is compared between the proposed algorithms with and without the edge-cutting method. The comparison is performed on a 3.4 GHz Quad-Core Intel(TM) i7 PC.
The results show that the edge-cutting method reduces the computation time effectively by rapidly achieving a solution with lower cost.
The message-passing algorithm, TWA, in \cite{bento2013message} is also tested to verify the effectiveness of the proposed method in our problem. TWA reduces the computation time compared to not applying it, but the amount is much smaller than the proposed method. 
Comparing Figs. \ref{fig:cutting_edge}(e) and (f), they verify that parallel computing further reduces the computation time effectively since our algorithm based on ADMM is capable of parallelization.

%	As shown in Figs. \ref{fig:cutting_edge}(d-f), parallel computing reduces the computation time, and if the $\epsilon$ value is properly set, the edge-cutting method can reduce the computation time while achieving the same result. 
%	The use of the message-passing algorithm in \cite{bento2013message} for our problem may reduce the computation time slightly, but the effect is less than that of the proposed edge-cutting method.

\begin{figure*}[t]
	\centering
	\subfigure[time = 32 sec.]{
		\includegraphics[width=.65\columnwidth]{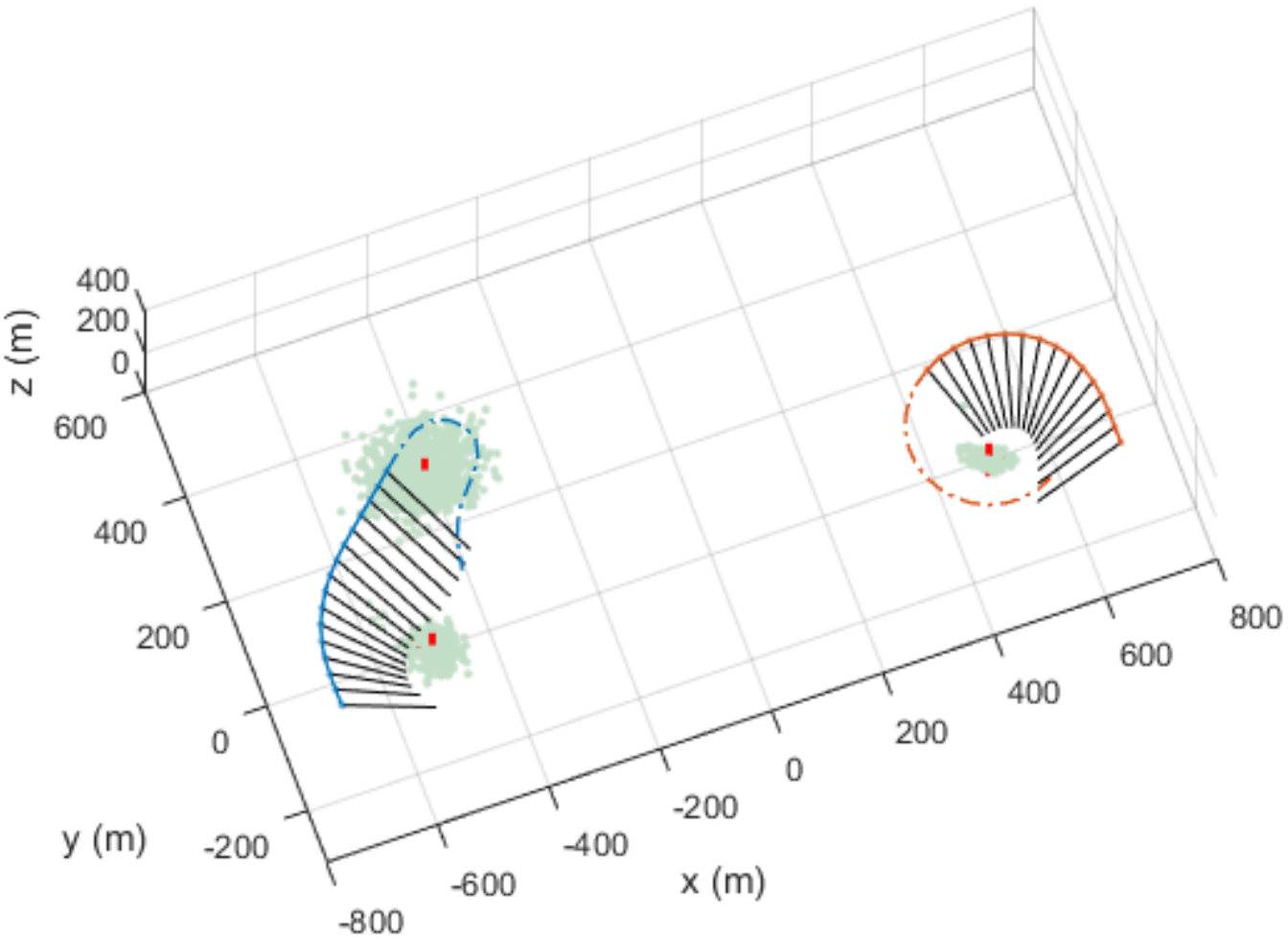}}
	\subfigure[time = 96 sec.]{
		\includegraphics[width=.65\columnwidth]{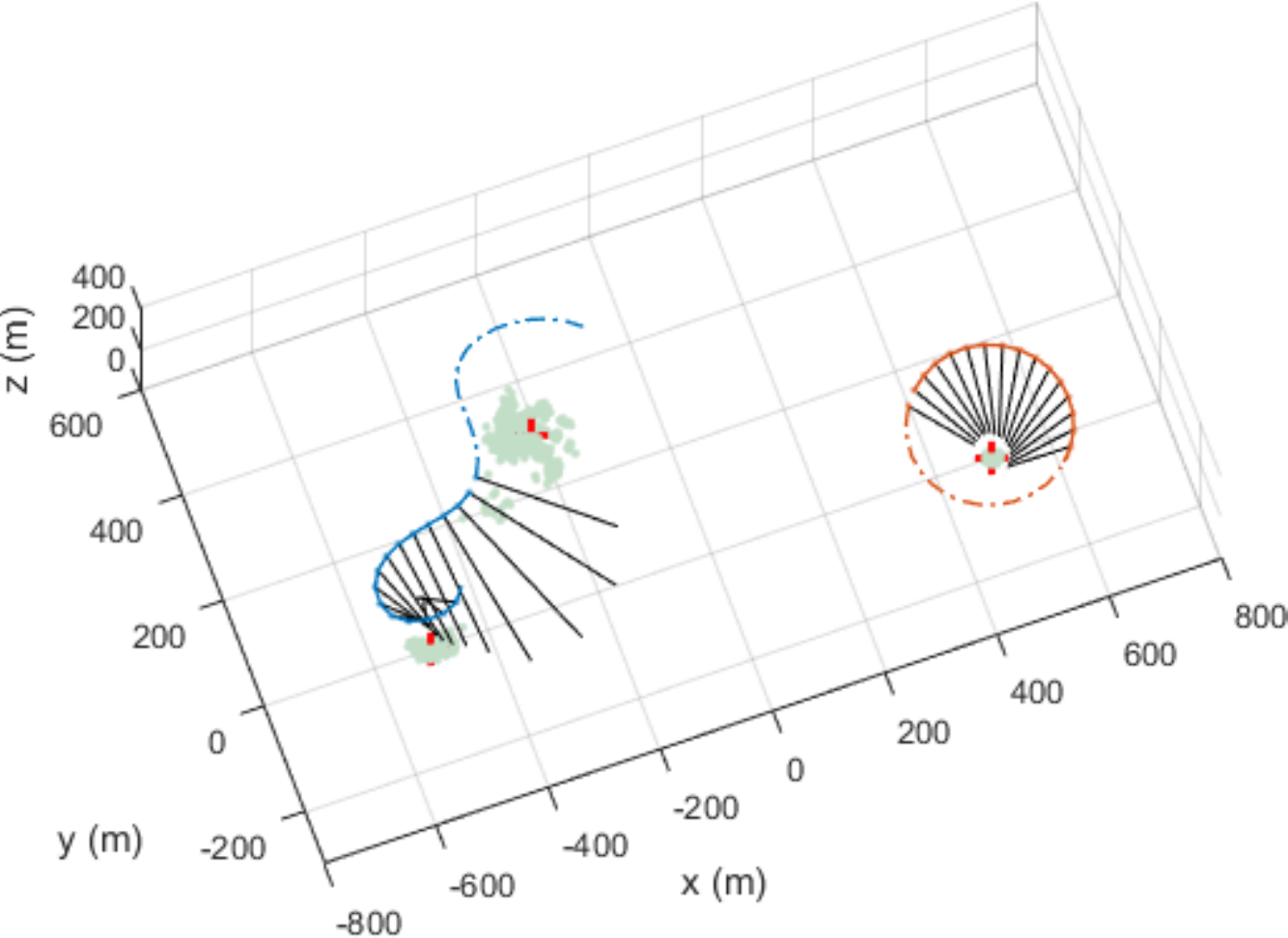}}
	\subfigure[time = 160 sec.]{
		\includegraphics[width=.65\columnwidth]{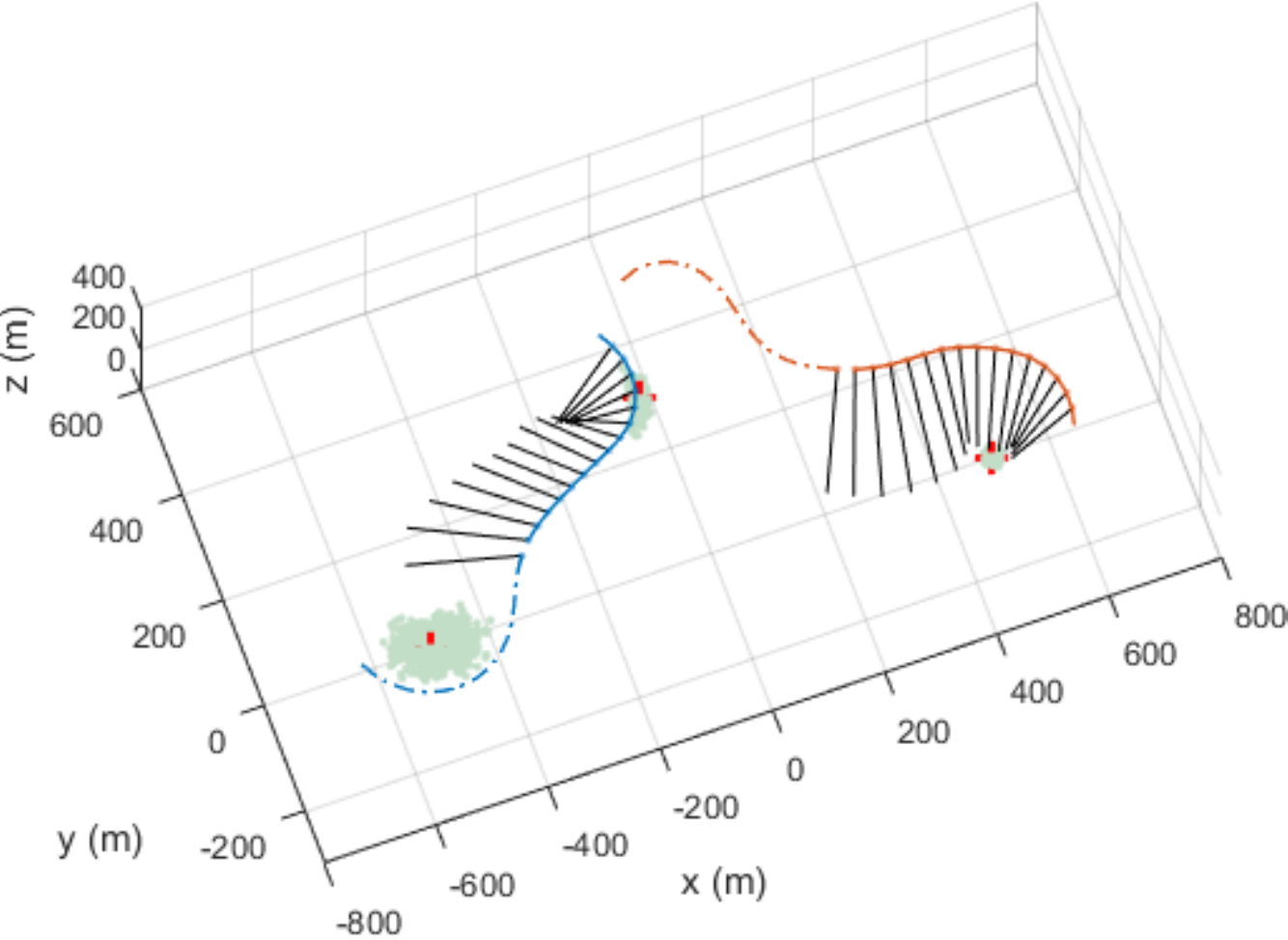}}
	\subfigure[time = 256 sec.]{
		\includegraphics[width=.65\columnwidth]{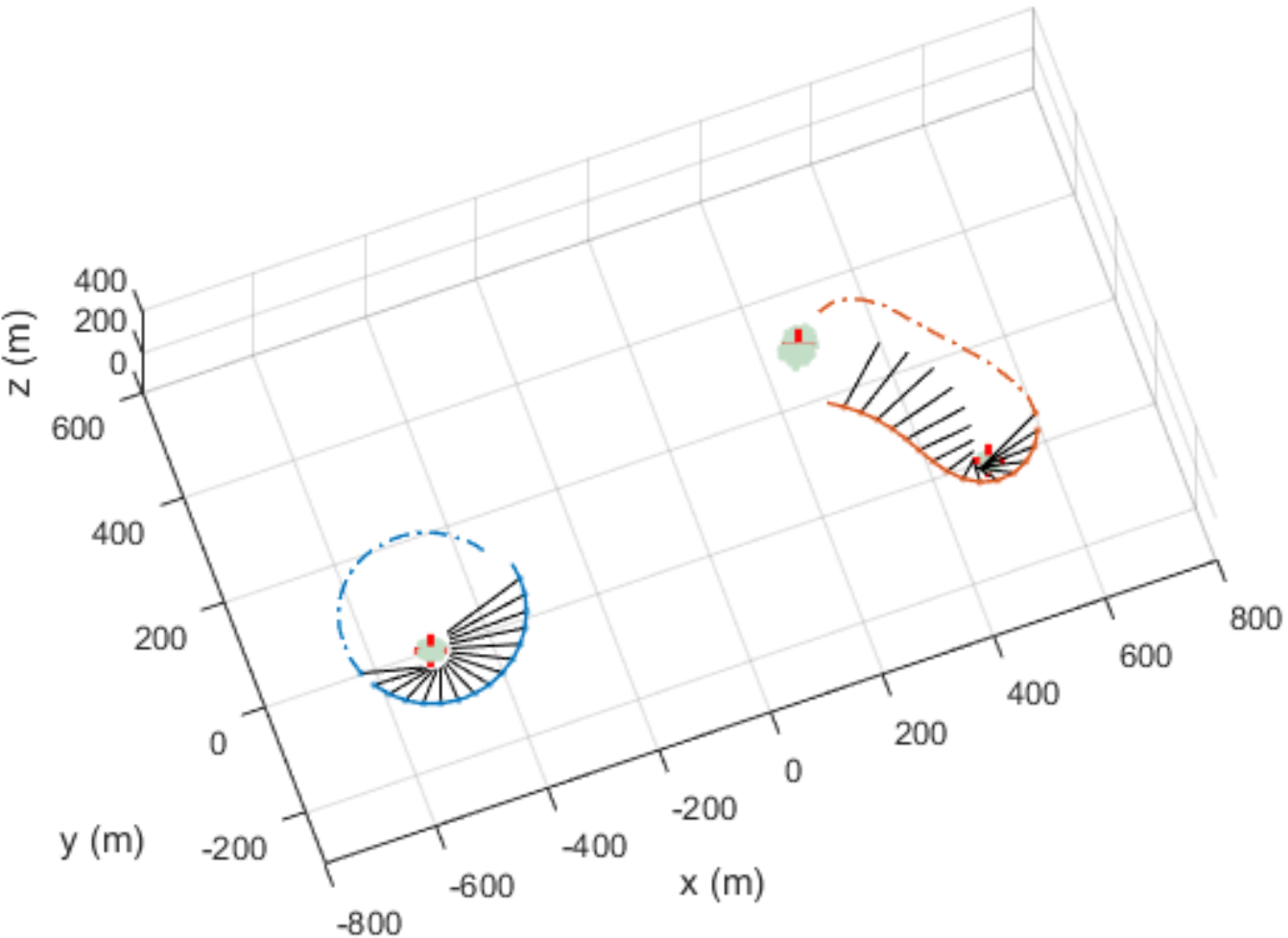}}
	\subfigure[Position error]{
		\includegraphics[width=.65\columnwidth]{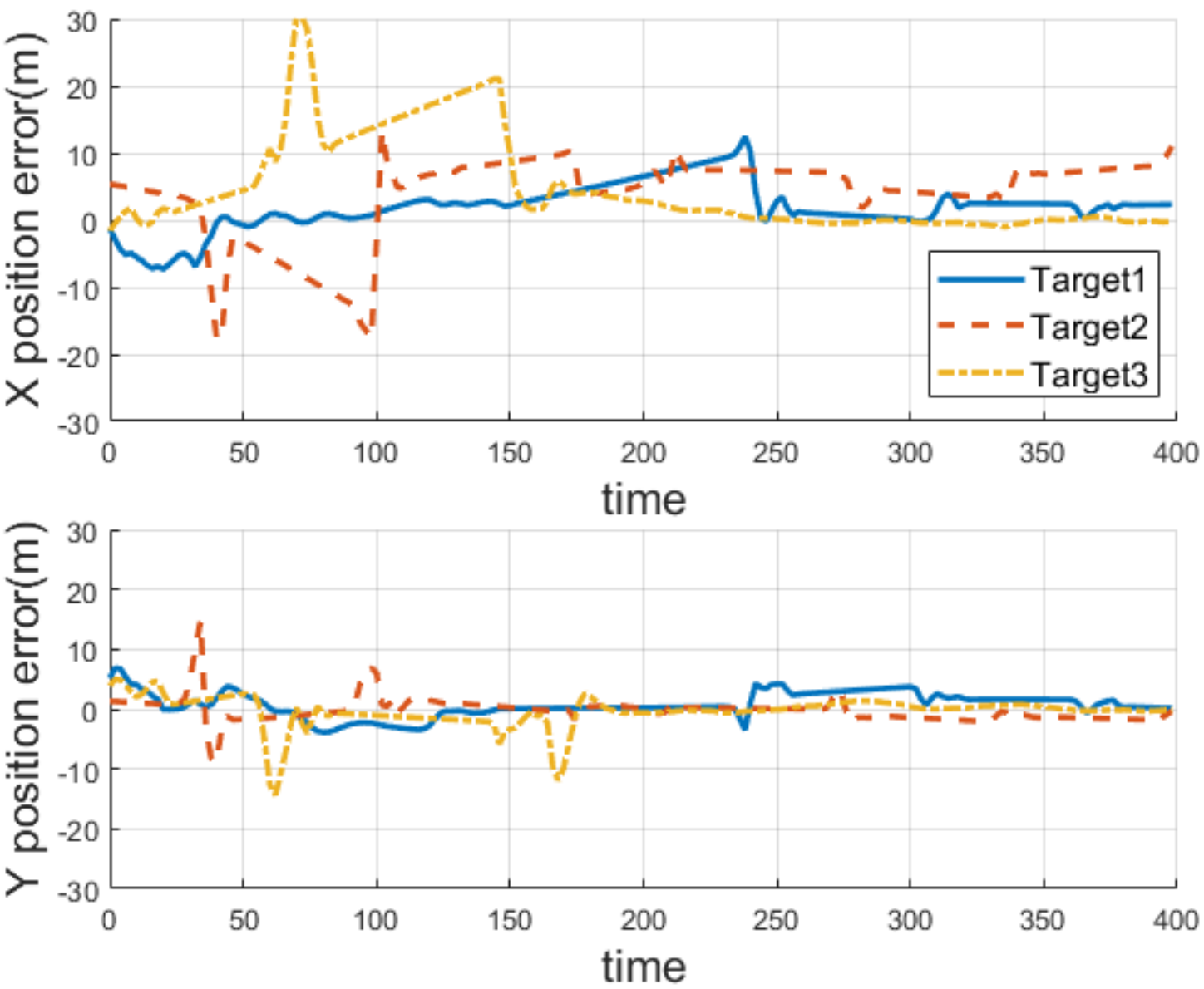}}
	\subfigure[Velocity error]{
		\includegraphics[width=.65\columnwidth]{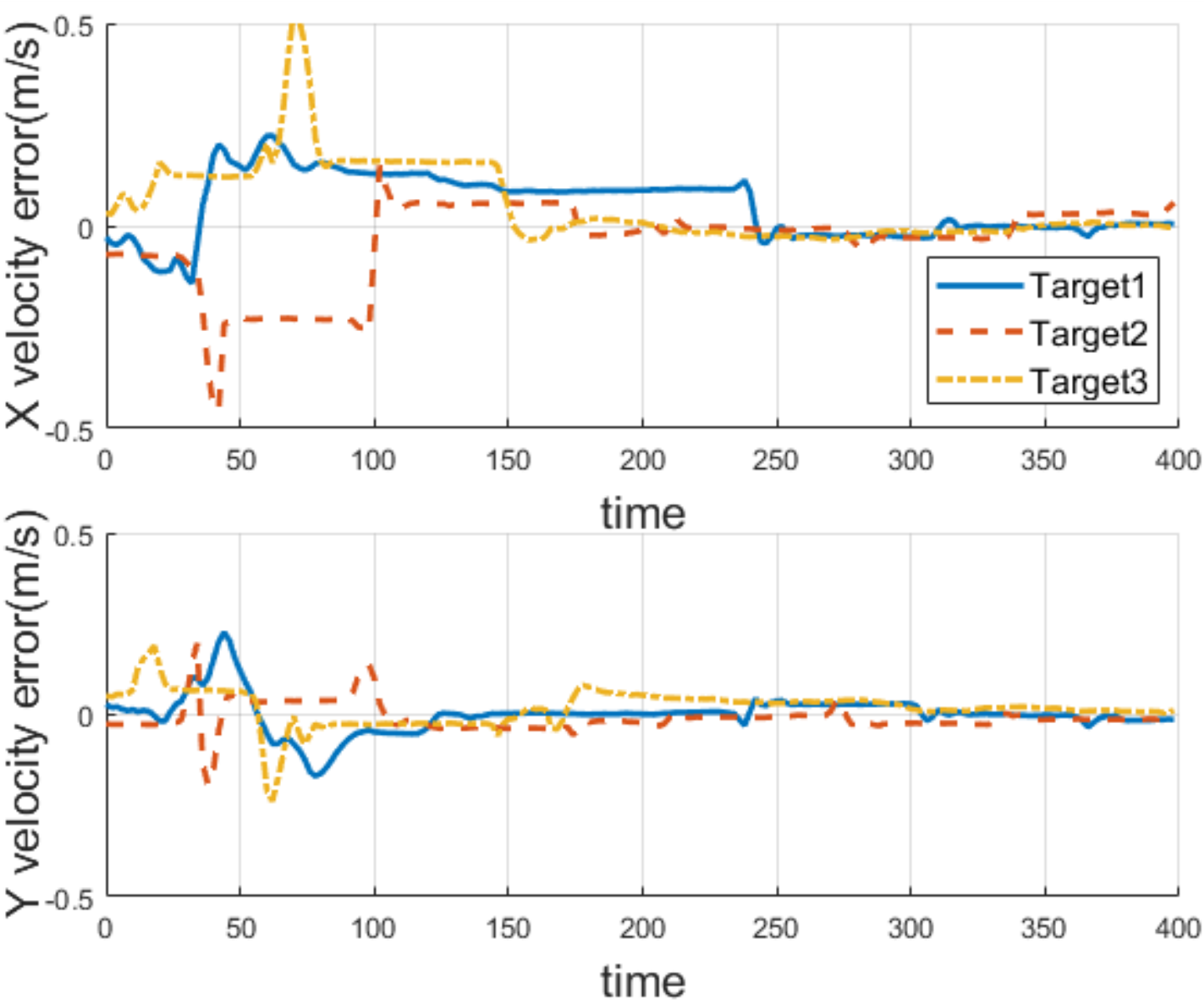}}
	\caption{(a)-(d) Snapshots of the current plan (solid line) and future plan (dashed line) for each mobile sensor in the replanning process scenario with two UAVs, one moving target and two stationary targets, and (e)-(f) particle filter estimation results}
	\label{fig: replanf}
\end{figure*} 

\subsection{Real-Time Operation Test}

In the last example, shown in Fig. \ref{fig: replanf}, we consider a scenario in which two UAVs track three targets, one with an initial velocity of 3m/s in the x-axis direction
and the others with zero initial velocities. We applied the distributed trajectory optimization algorithm, the edge-cutting method, and the modified RHC scheme all together. It is assumed that a particle filter is used to estimate the states of the targets in the actual operating environment of the UAVs. 
The control horizon, $T_c$, is set to 25 seconds considering the computation time of the algorithms. It is assumed that the sensors are attached to the left side of one UAV and to the right side of the other UAV, and that UAVs fly at different altitudes of $100m$ and $120m$, respectively.
Figs. \ref{fig: replanf}(a-d) show some of the simulated results under the assumption that the UAVs operate for 400 seconds. 
A notable point in this example is that the UAV automatically passes the tracking mission to the other UAV after about 160 seconds.  
Figs. \ref{fig: replanf}(e-f) show the state estimation results of the targets, where it can be confirmed that all targets are tracked reliably through the modified RHC scheme.

\section{Conclusions}

In this paper, we investigated a distributed optimization approach to trajectory planning for a multi-target tracking problem. In order to solve the trajectory optimization and task-assignment problems simultaneously, the distributed trajectory optimization problem was formulated, and then solved by integrating a variant of a differential dynamic programming algorithm called iterative Linear-Quadratic-Gaussian (iLQG) algorithm with the distributed Alternating Direction Method of Multipliers (ADMM). In addition, we proposed an edge-cutting method to reduce the computation time of the algorithm and the modified RHC scheme for real-time operation. 
Numerical experiments were conducted and the results presented to demonstrate the applicability and validity of the proposed approach.

%\section*{Acknowledgment}
%This work was conducted at High-Speed Vehicle Research Center of KAIST with the support of the Defense Acquisition Program Administration and the Agency for Defense Development under Contract UD170018CD. 

	\addtolength{\textheight}{-1cm}   % This command serves to balance the column lengths
	% on the last page of the document manually. It shortens
	% the textheight of the last page by a suitable amount.
	% This command does not take effect until the next page
	% so it should come on the page before the last. Make
	% sure that you do not shorten the textheight too much.
	
	%%%%%%%%%%%%%%%%%%%%%%%%%%%%%%%%%%%%%%%%%%%%%%%%%%%%%%%%%%%%%%%%%%%%%%%%%%%%%%%%

	%%%%%%%%%%%%%%%%%%%%%%%%%%%%%%%%%%%%%%%%%%%%%%%%%%%%%%%%%%%%%%%%%%%%%%%%%%%%%%%%

	%%%%%%%%%%%%%%%%%%%%%%%%%%%%%%%%%%%%%%%%%%%%%%%%%%%%%%%%%%%%%%%%%%%%%%%%%%%%%%%%

	\bibliographystyle{IEEEtran}
	\bibliography{manuscript}
	
\end{document}